\documentclass{article}

\usepackage[main, final]{neurips_2026}

\usepackage[utf8]{inputenc} 
\usepackage[T1]{fontenc}    
\usepackage{hyperref}       
\usepackage{url}            
\usepackage{booktabs}       
\usepackage{amsfonts}       
\usepackage{nicefrac}       
\usepackage{microtype}      
\usepackage{xcolor}         
\usepackage{wrapfig}
\usepackage{amsmath} 
\usepackage{graphicx} 

\title{NP-LoRA: Null Space Projection for Subject-Style LoRA Fusion}

\author{%
  \makebox[\textwidth][c]{%
    \begin{minipage}[t]{0.32\textwidth}\centering
      \textbf{Chuheng Chen}\textsuperscript{1,2}\\[-1pt]
      \texttt{chenchuheng@iie.ac.cn}
    \end{minipage}\hfill
    \begin{minipage}[t]{0.32\textwidth}\centering
      \textbf{Xiaofei Zhou}\textsuperscript{1,2}\thanks{Corresponding author.}\\[-1pt]
      \texttt{zhouxiaofei@iie.ac.cn}
    \end{minipage}\hfill
    \begin{minipage}[t]{0.32\textwidth}\centering
      \textbf{Geyuan Zhang}\textsuperscript{1,2}\\[-1pt]
      \texttt{zhanggeyuan@iie.ac.cn}
    \end{minipage}%
  }\\[1.8em]
  \makebox[\textwidth][c]{%
    \begin{minipage}[t]{0.32\textwidth}\centering
      \textbf{Yong Huang}\textsuperscript{1,2}\\[-1pt]
      \texttt{huangyong@iie.ac.cn}
    \end{minipage}%
    \begin{minipage}[t]{0.32\textwidth}\centering
      \textbf{Gang Xiong}\textsuperscript{1,2}\\[-1pt]
      \texttt{xionggang@iie.ac.cn}
    \end{minipage}%
  }\\[1.8em]
  \textsuperscript{1}Institute of Information Engineering, Chinese Academy of Sciences, Beijing, China \\
  \textsuperscript{2}School of Cyber Security, University of Chinese Academy of Sciences, Beijing, China
}

\begin{document}

\maketitle

\begin{abstract}
Low-Rank Adaptation (LoRA) fusion enables the composition of subject and style representations for controllable generation without retraining. 
However, existing approaches primarily operate through weight-level merging, without explicitly modeling how independently trained LoRAs interact in the shared parameter space. 
We adopt a geometric perspective on LoRA fusion, interpreting content and style LoRAs as occupying overlapping, non-orthogonal low-rank subspaces, where such overlap can lead to conflicting parameter updates that affect generation quality. 
This observation motivates us to reformulate LoRA fusion not merely as parameter combination, but as a problem of controlling how updates from overlapping subspaces are combined. 
Based on this insight, we propose Null Space Projection LoRA (NP-LoRA), a training-free framework that employs projection as a fusion operator to explicitly modulate cross-LoRA interactions. 
Specifically, NP-LoRA uses principal directions of the style LoRA to define a projection subspace and projects the content LoRA onto the complementary subspace (i.e., the null space of the style LoRA), suppressing interference along dominant style directions while preserving complementary information.
To avoid the overly aggressive suppression of hard projection, we further formulate soft projection as a regularized optimization problem that balances content preservation against style-subspace suppression. 
This objective admits a closed-form solution, yielding a projection operator controlled by a single parameter that continuously interpolates between linear merging and hard projection. 
Extensive experiments across multiple pretrained LoRA pairs show that NP-LoRA achieves more balanced content-style composition compared to strong baselines, without requiring retraining.
Code is available at https://github.com/blackzebra7/NP-LoRA.
\end{abstract}



\begin{wrapfigure}[35]{R}{0.52\textwidth}  
\vspace{-4pt} 
  \centering
  \includegraphics[width=\linewidth, trim=10bp 10bp 10bp 10bp, clip]{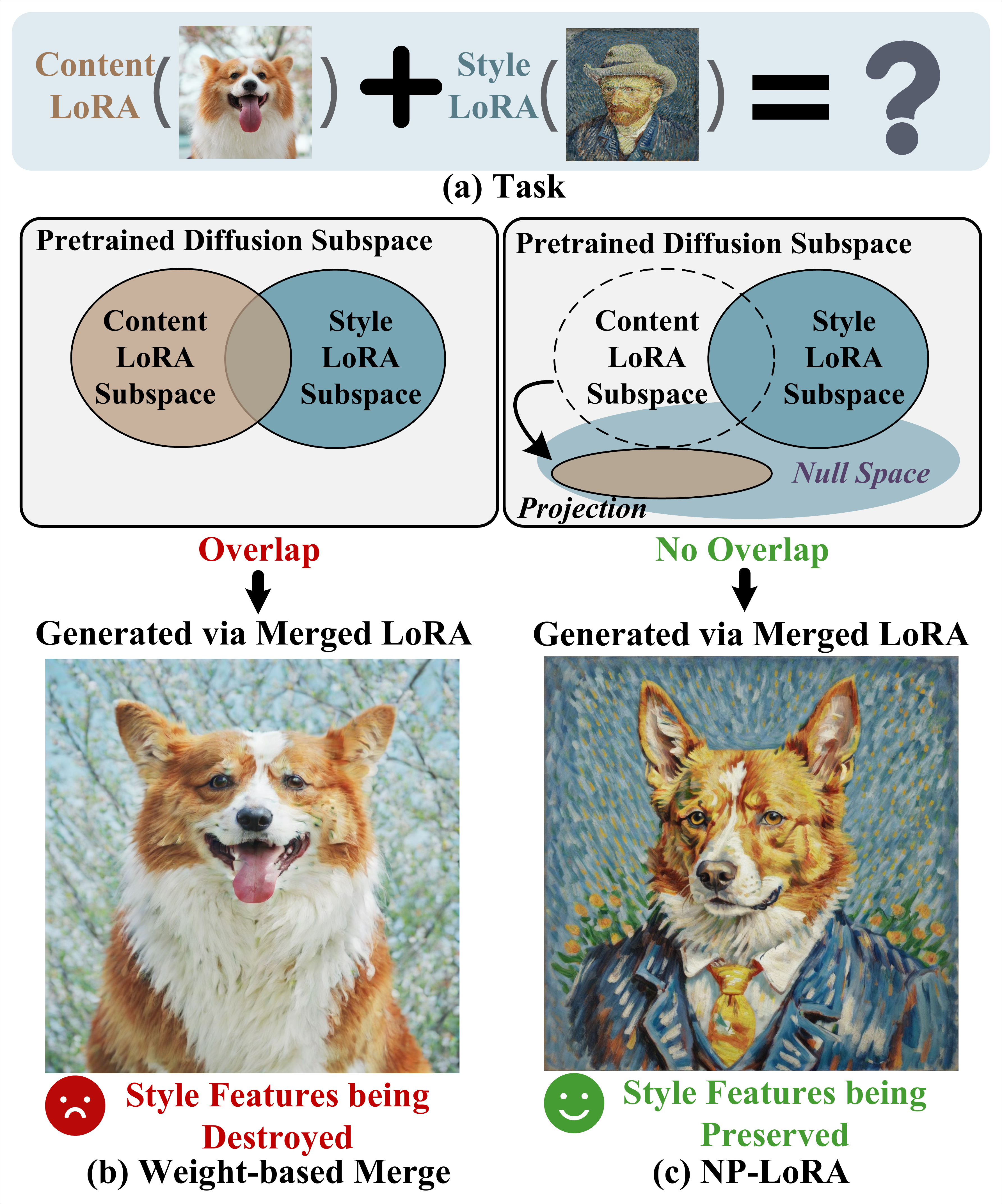}
  \caption{
    Illustration of our motivation. 
    (a) The task is to compose a content LoRA (subject identity) with a style LoRA (artistic appearance) for controllable generation.
    (b) Weighted merging can degrade stylistic fidelity, potentially due to interference, as independently trained LoRAs may occupy correlated, non-orthogonal low-rank subspaces of the diffusion feature space.
    (c) NP-LoRA projects the content LoRA into the null space of the style LoRA, reducing overlap and better preserving stylistic characteristics.
  }
  \label{fig:introduction_hard}
\end{wrapfigure}

\section{introduction}
Low-Rank Adaptation (LoRA)~\cite{hu2022lora} has become a standard approach for customizing diffusion models~\cite{ddpm, ddim_1, ddim_2, ldm}, enabling subject- or style-specific adaptation from only a few examples~\cite{dreambooth}.
In practice, users typically train LoRAs independently for different factors, such as subject identity and artistic style, due to differences in data, prompts, and objectives.
This naturally leads to a common requirement: composing multiple independently trained LoRAs for flexible and reusable generation.
For example, users may wish to render a learned subject (e.g., a pet dog) in a separately learned artistic style (e.g., Van Gogh’s brushwork), as illustrated in Fig.~\ref{fig:introduction_hard}(a).
However, such fusion is often unreliable in practice: naive merging frequently produces degraded results, where either subject identity is distorted or the intended style is weakened (Fig.~\ref{fig:introduction_hard}(b)).
This behavior highlights a key challenge in personalized concept composition, where independently trained LoRAs must be combined while preserving their respective functionalities.

This difficulty arises because independently trained LoRAs share the same underlying model space, causing their parameter updates to compete along shared parameter directions during fusion.
Existing methods primarily operate through weight-level merging in parameter space.
Recent approaches, such as ZipLoRA~\cite{ziplora}, DuoLoRA~\cite{roy2025duolora}, K-LoRA~\cite{klora}, and LoRA.rar~\cite{lorarar}, improve over naive addition through optimized fusion weights, timestep-dependent switching, or learned hypernetworks.
While effective, these methods rely on heuristic combinations of LoRA updates and do not explicitly model how different LoRAs interact, often resulting in trade-offs between subject fidelity and stylistic consistency.
We interpret these failure modes from a geometric perspective.
Since LoRAs are trained within the same model space, content and style LoRAs may occupy correlated low-rank subspaces, leading to overlapping directions that may encode inconsistent semantics.
Importantly, we do not assume that LoRAs are cleanly disentangled; instead, we consider realistic settings where semantics partially overlap.
When updates from different LoRAs compete along these shared directions, the merged model may become biased toward one concept, degrading the balance between subject identity and style consistency.
The goal is therefore to combine the content LoRA ($\Delta W_c$) and the style LoRA ($\Delta W_s$) into a single module that preserves both representations while mitigating such interference.
We focus on pairwise subject--style composition, where the two LoRAs have distinct roles in preserving subject identity and transferring visual style.

Our empirical analysis (Sec.~\ref{sec:exp_perturb}) reveals that LoRAs exhibit anisotropic low-rank structures, where a small number of dominant directions contribute disproportionately to generation and are particularly sensitive to cross-LoRA interference.
This suggests that interference is concentrated along a few critical directions, making simple weight-level combinations insufficient.
Consequently, effective fusion requires direction-wise control, selectively suppressing interfering components while preserving complementary ones.
Motivated by this, we move beyond weight-level merging and instead explicitly control interactions in the shared subspace via projection.

To this end, we propose Null Space Projection LoRA (NP-LoRA), a training-free framework that redefines projection as a fusion operator for composing independently trained LoRAs. 
Unlike existing methods that rely on heuristic merging, NP-LoRA explicitly controls cross-LoRA interactions by modulating contributions along shared subspace directions. 
Specifically, we perform singular value decomposition (SVD) on the style LoRA $\Delta W_s$ to extract its dominant directions, and project the content LoRA $\Delta W_c$ onto the complementary subspace, suppressing components aligned with style-sensitive directions while preserving complementary information. 
To avoid the overly aggressive behavior of hard projection, we further formulate soft projection as a regularized optimization problem that balances content preservation and style-subspace suppression. 
This objective admits a closed-form solution, yielding a projection operator controlled by a single parameter that continuously interpolates between linear merging and hard projection. 
NP-LoRA is training-free, requiring no additional data or optimization, and can be directly applied to pretrained LoRAs in a plug-and-play manner. 
This makes it particularly suitable for practical scenarios where users wish to compose LoRAs efficiently without retraining. 
Extensive experiments show that it achieves more balanced content--style composition than strong baselines across diverse settings.

Our main contributions are summarized as follows:
\begin{itemize}
\item We provide a geometric perspective on LoRA fusion, observing that independently trained LoRAs can exhibit substantial overlap in their low-rank subspaces, which may contribute to interference during fusion. This perspective offers an intuitive explanation for common failure patterns in existing merging methods.
\item We propose Null Space Projection LoRA (NP-LoRA), a novel training-free framework that reformulates LoRA fusion as controlling contributions along shared subspace directions, and introduces projection as an explicit fusion operator for modulating cross-LoRA interactions.
\item We formulate soft projection as a regularized optimization problem with a closed-form solution, yielding a parameterized operator that interpolates between linear merging and hard projection, enabling flexible control over the content--style trade-off.
\item We conduct extensive experiments across multiple backbones, LoRA ranks, and pretrained LoRA pairs, demonstrating that NP-LoRA consistently achieves more balanced and robust content--style composition compared to strong baselines.
\end{itemize}

\section{Related Work}
\noindent\textbf{Customization and LoRA composition in diffusion models.}
Diffusion models~\cite{ddpm, ddim_1, ddim_2, ldm} have supported image generation and broader visual tasks, including image restoration and generative refinement for concealed visual perception~\cite{he2025retidiff,he2025reversible}.
In text-to-image customization, methods range from early few-shot personalization~\cite{textinver_1, textinver_2, textinver_3, dreambooth} to efficient, reference-conditioned, and unified subject--style generation~\cite{customdiffusion, avrahami2023break, shi2024instantbooth, xiao2025fastcomposer, xie2023smartbrush,11206511,wu2026unified,xu2026unicustom}. 
Low-Rank Adaptation (LoRA)~\cite{hu2022lora} and its variants~\cite{hayou2024lora+, kopiczko2023vera, ren2024melora, zhang2023lora, zhou2024lora, zi2023delta} provide efficient alternatives via rank-decomposed adapters, enabling modular customization. 
Prior work has explored LoRA composition by learning more disentangled representations (e.g., B-LoRA~\cite{blora}, B4M~\cite{xu2025b4m}) or by designing mechanisms for multi-concept generation (e.g., CSGO~\cite{xing2024csgo}). 
In contrast, we focus on \emph{post-hoc composition of independently trained LoRAs}, where no retraining is assumed.
Existing LoRA composition methods for image generation can be roughly categorized into object composition~\cite{dong2024continually, gu2023mix, jiang2025mc, liu2023cones, yang2024lora, zhong2024multi, zhang2025rethinking, zou2025cached, simsar2025loraclr} and content-style fusion~\cite{ziplora, klora, lorarar, estlora}. 
Recent methods compose multiple LoRAs through contrastive test-time optimization, direct composition, intrinsic skill composition, cached composition, or a learned weight basis~\cite{11444429, zhong2024multilora, foteinopoulou2025loratoriointrinsicapproachlora, ICLR2025_bc8f76d9, manor2026lorweb}. 
Recent content-style methods, such as ZipLoRA~\cite{ziplora}, DuoLoRA~\cite{roy2025duolora}, K-LoRA~\cite{klora}, and LoRA.rar~\cite{lorarar}, improve over naive addition through weighted fusion, adaptive switching, or hypernetworks. 
However, these approaches operate primarily through weight-level operations and do not explicitly model the geometry of cross-LoRA subspace interactions. 
Our work is complementary to these strategies: instead of designing new training procedures or learned fusion modules, we analyze post-hoc LoRA fusion through the geometry of low-rank subspaces and introduce a projection-based operator to control their interactions.

\noindent\textbf{Orthogonality and Subspace Methods in Model Adaptation.}
Orthogonality and subspace methods are widely used in model adaptation to mitigate interference, often through singular value decomposition (SVD), null-space constraints, or orthogonal projections.
Recent works~\cite{knots, subzero, prolora, fangalphaedit, zhang2025lion, xiong2026oplora} leverage these ideas in different settings, but target objectives distinct from post-hoc LoRA fusion.
KnOTS~\cite{knots} uses SVD to extract shared bases for LoRA alignment and aggregation, while SubZero~\cite{subzero} reduces content--style leakage through orthogonal cross-attention at the feature level.
LiON-LoRA~\cite{zhang2025lion} focuses on coordinating spatial--temporal LoRAs in video diffusion.
Closest to our work, ProLoRA~\cite{prolora} and LoRA-Null~\cite{tang2025lora} also exploit orthogonal or null-space properties across LoRAs, but primarily use them for re-adaptation, initialization, or interference reduction during sequential learning.
In contrast, we study \emph{training-free post-hoc composition of independently trained content and style LoRAs}.
Rather than using projection as a training constraint or initialization strategy, we use it directly as a fusion operator and derive a closed-form soft projection that balances content preservation and style-subspace suppression.

\section{Background}

\noindent\textbf{LoRA Fine-tuning.} 
Low-Rank Adaptation (LoRA)~\cite{hu2022lora} efficiently adapts pretrained diffusion models~\cite{ddpm, ddim_1, ddim_2, ldm} by inserting low-rank updates into existing weight matrices. 
Given a pretrained weight $W_0 \in \mathbb{R}^{m \times n}$, LoRA parameterizes the update as
\begin{equation}
\Delta W = BA, \quad B \in \mathbb{R}^{m\times r}, \; A \in \mathbb{R}^{r\times n}, \; r \ll \min(m,n).
\end{equation}
During training, only $A$ and $B$ are optimized while $W_0$ is frozen. 
At inference, the adapted weight becomes $W = W_0 + \Delta W$.

\noindent\textbf{Problem Setup.} 
We consider the problem of composing two independently trained LoRAs: a content LoRA $\Delta W_c$ capturing subject identity, and a style LoRA $\Delta W_s$ capturing visual style. 
Given a pretrained diffusion model, our goal is to construct a merged LoRA $\Delta W_m$:
\begin{equation}
\Delta W_m = \text{Merge}(\Delta W_c, \Delta W_s),
\end{equation}
such that both content and style are preserved during generation. 
We omit layer indices for clarity, as the same operation is applied to each LoRA matrix~\cite{ziplora}. 
This setup focuses on post-hoc composition, where LoRAs are trained independently and must be combined without retraining.

\begin{figure*}[t]
    \centering
    \includegraphics[width=\linewidth, trim=10bp 10bp 10bp 10bp, clip]{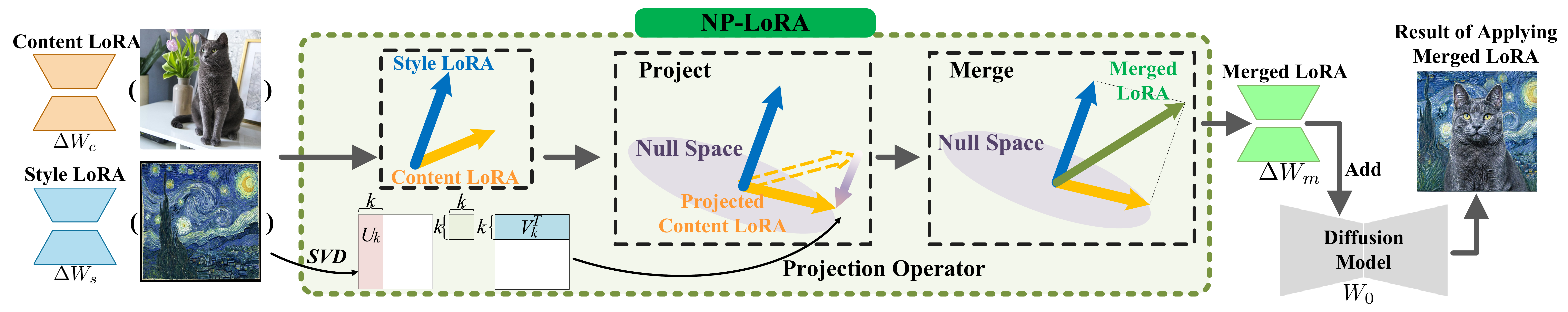}
    \caption{Overview of the proposed method.
    NP-LoRA takes pretrained content and style LoRAs as inputs.
    The style LoRA is decomposed via singular value decomposition (SVD) to construct a null space, onto which the content LoRA is projected.
    This design enables training-free fusion through a closed-form projection operator.}
    \label{fig:method_overview}
\end{figure*}

\section{Methodology}
In this work, we propose NP-LoRA, a training-free framework for fusing independently trained LoRAs (Fig.~\ref{fig:method_overview}).
We view LoRA fusion as controlling interactions between overlapping subspaces.
We first apply singular value decomposition (SVD) to identify dominant directions of the style LoRA, and project the content LoRA onto the complementary subspace (i.e., the null space of the style LoRA).
This suppresses its contribution along style-critical directions while retaining complementary information.
In the limiting case, a hard projection strongly reduces overlap but may discard useful components.
We therefore introduce a soft projection with a tunable parameter, enabling continuous control over the trade-off between subject fidelity and style consistency.

\subsection{Principal Direction Extraction via SVD} \label{sec:exp_perturb}
We analyze LoRA representations through singular value decomposition (SVD), decomposing the style LoRA as
\begin{equation}
\Delta W_s = U \Sigma V^\top,
\end{equation}
where the columns of $V \in \mathbb{R}^{d_{in} \times r_s}$ are orthonormal right singular vectors.
Each vector $v \in V$ defines a direction in parameter space, and its corresponding singular value in $\Sigma$ reflects the strength of $\Delta W_s$ along that direction.

\begin{wrapfigure}[20]{r}{0.50\textwidth}
    \vspace{-16pt} 
    \centering
    \includegraphics[width=0.5\textwidth, trim=5bp 5bp 5bp 5bp, clip]{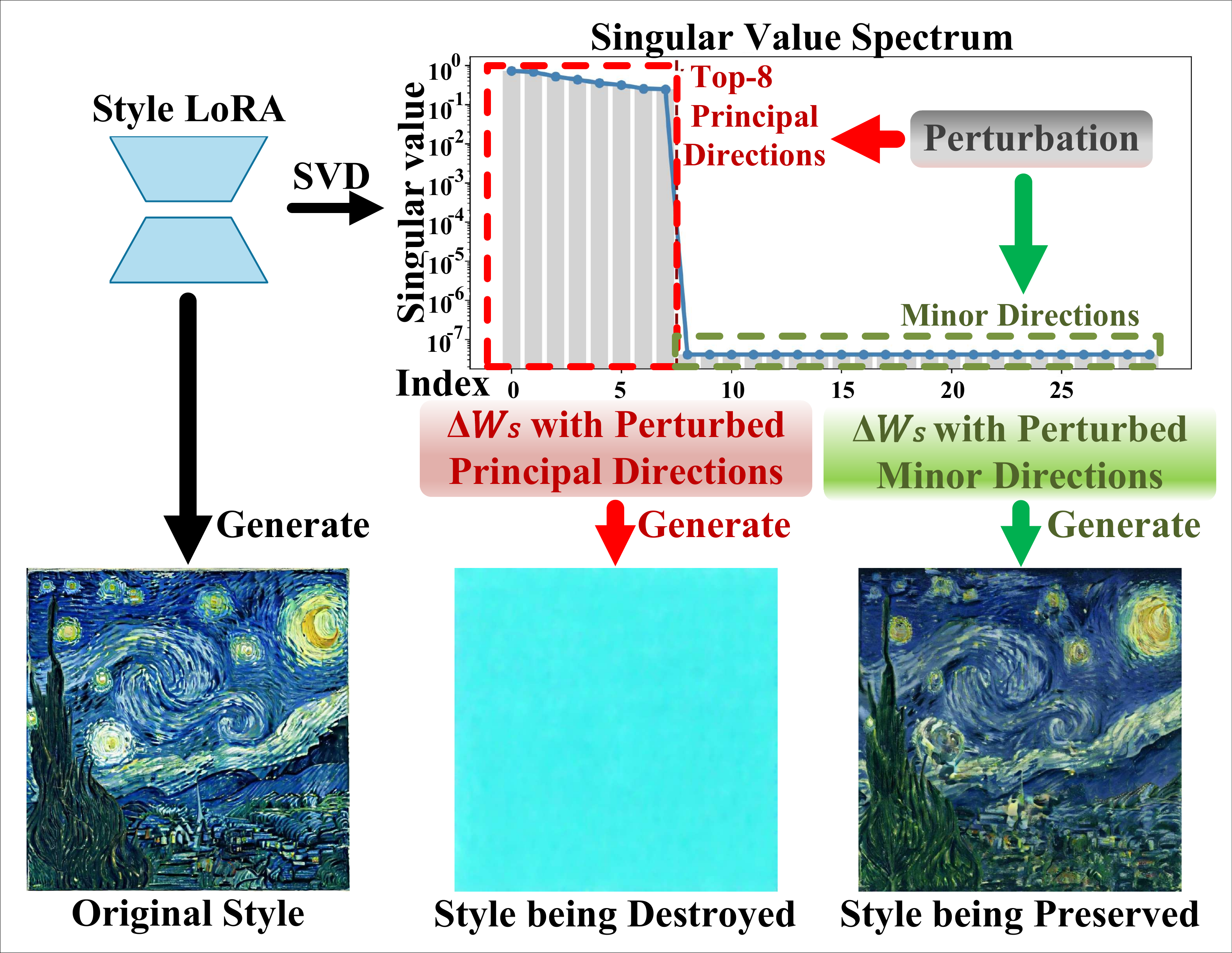}
    \caption{Singular value spectrum of a LoRA and perturbation effects. Perturbing dominant directions produces more noticeable stylistic changes than perturbing less principal directions.}
    \label{fig:style-subspace}
\end{wrapfigure}

Following common practice~\cite{klora, dreambooth}, we treat the leading singular vectors as dominant directions in the LoRA update space. As shown in Fig.~\ref{fig:style-subspace}, perturbing dominant directions produces more noticeable stylistic changes than less dominant ones, suggesting that certain directions may contribute more strongly to generation and be more susceptible to interference during LoRA fusion. We emphasize that this analysis is intended as an illustrative geometric observation rather than a rigorous statistical characterization of LoRA interference.

Having identified these dominant style directions, we next examine how they behave under existing merging strategies. 
A common approach is linear fusion:
\begin{equation} \label{Eq:linear_weight} 
\Delta W_{\mathrm{m}} = a\Delta W_c + b\Delta W_s, \quad a,b \in \mathbb{R}.
\end{equation}
However, such combinations do not explicitly account for interactions between LoRAs, and may introduce competing contributions along shared directions, potentially degrading stylistic consistency.

\paragraph{Why weight-based merging fails.}
We analyze whether weighted combinations can preserve style-critical directions. 
Let the style subspace be spanned by the dominant singular vectors of $\Delta W_s$, with projection operator $P = V_k V_k^\top$. 

We decompose the content LoRA as
\begin{equation}
\Delta W_c = \Delta W_c (I - P) + \Delta W_c P,
\end{equation}
where $\Delta W_c P$ lies in the style subspace.

For a weighted merge $\Delta W_m = a\Delta W_c + b\Delta W_s$, its projection onto the style subspace is
\begin{equation}
\Delta W_m P = a\,\Delta W_c P + b\,\Delta W_s.
\end{equation}

For nonzero $a$, preserving the original style component would require $a\,\Delta W_c P \approx (1-b)\,\Delta W_s$.
Because the same coefficient $a$ scales content contributions both within and outside the style subspace, weighted merging cannot adjust the two independently.



\subsection{Hard Projection for Subspace Control}

Having identified the dominant style directions and observed that weight-based merging does not explicitly account for their interactions, we introduce a projection-based formulation to modulate cross-LoRA contributions. 
We use the style LoRA as the reference because, in our evaluated subject--style pairs, direct merging tends to weaken stylistic characteristics more than subject structure.

We define the projection onto the dominant style subspace as $P = V_k V_k^\top$, where $V_k$ spans the style subspace.
In practice, $k$ is set to the LoRA rank, so that the full low-rank subspace is considered without introducing additional selection hyperparameters.
The null space projection is given by:
\begin{equation} \label{eq:hard_projection}
    P_{\text{null}} = I - P = I - V_k V_k^\top,
\end{equation}
which suppresses components aligned with style-critical directions while retaining complementary information.

Applying this projection to the content LoRA $\Delta W_c \in \mathbb{R}^{m \times n}$ yields:
\begin{equation} \label{eq12}
    \Delta W_c^{\text{proj}} = \Delta W_c P_{\text{null}} = \Delta W_c (I - V_k V_k^\top).
\end{equation}
The merged LoRA is then defined as:
\begin{align}\label{eq:hard}
    \Delta W_m 
    &= \Delta W_s + \Delta W_c^{\text{proj}} \\
    &= \Delta W_s + \Delta W_c (I - V_k V_k^\top). \nonumber
\end{align}

This formulation restricts the contribution of the content LoRA along dominant style directions, allowing the style LoRA $\Delta W_s$ to primarily govern these subspaces.

From a geometric perspective, this can be viewed as reducing the overlap between content and style subspaces along critical directions. 
In the limiting case, this projection minimizes the contribution of $\Delta W_c$ within the style subspace, which can be quantified by:
\begin{equation}
\|\Delta W_c^{\text{proj}} P\|_F^2,
\end{equation}
representing the residual energy of the projected content LoRA within the style subspace.

\subsection{Soft Projection for Balanced Fusion}

Instead of treating projection as a binary operation, we reinterpret LoRA fusion as a continuous subspace interaction problem. 
In practice, content and style LoRAs often share partially aligned directions, and hard projection may overly suppress useful components. 
This motivates a flexible mechanism that modulates their interaction rather than strictly eliminating overlap.

We introduce a subspace-aware formulation:
\begin{equation}
    \min_{\Delta W_c^{\text{proj}}}\; 
    \|\Delta W_c^{\text{proj}} - \Delta W_c\|_F^2 
    + \mu\,\|\Delta W_c^{\text{proj}} P\|_F^2,
\end{equation}
where the first term preserves content, and the second suppresses its energy within the style subspace. 
The parameter $\mu \ge 0$ controls the strength of this subspace interaction.

This optimization admits a closed-form solution, which leads directly to the merged LoRA:
\begin{equation} \label{eq:soft_proj}
    \Delta W_m 
    = \Delta W_s + \Delta W_c (I+\mu P)^{-1}
    = \Delta W_s + \Delta W_c \Big(I - \tfrac{\mu}{1+\mu}V_k V_k^\top\Big).
\end{equation}

Equivalently, $\mu$ removes a fraction $\mu/(1+\mu)$ of the content component within the style subspace, retaining the remaining $1/(1+\mu)$, while leaving its complementary component unchanged; Tab.~\ref{tab:mu_sensitivity} shows the resulting trade-off.

We apply this modulation to the content LoRA while using the style subspace as the reference for controlling their interaction. 
As $\mu \to 0$, the formulation reduces to linear merging, while $\mu \to \infty$ approaches hard projection. 
Thus, $\mu$ provides continuous control over the trade-off between content fidelity and style preservation.
Unlike strict orthogonal projection, our formulation continuously attenuates overlapping content and style components while preserving complementary ones. This allows effective fusion when the LoRA subspaces are partially aligned, without requiring complete separation.

\section{Experiments}
\subsection{Experiment Setup}

\textbf{Datasets.}
Following common practice in customization~\cite{ziplora,klora}, we obtain LoRAs from publicly available images.  
For content, we adopt the DreamBooth~\cite{ruiz2023dreambooth} dataset, where each subject is represented by 4--5 reference images.  
For style, we use the StyleDrop~\cite{styledrop} dataset, complemented with several representative artistic styles.  
For community-trained LoRAs, we use publicly available models from Hugging Face for evaluation as shown in Appendix.  


\noindent\textbf{Experimental details.}  
The main experiments use SDXL v1.0~\cite{sdxl}; the FLUX evaluation is reported separately.
For training the LoRAs to be merged, including the training prompts and ranks, we follow the settings in K-LoRA~\cite{klora, dreambooth}.
Importantly, NP-LoRA does not assume clean disentanglement or orthogonality between content and style LoRAs, but instead is designed to handle their interactions under potentially overlapping and entangled representations.
We use a global $\mu=0.5$ in the soft projection (Eq.~\ref{eq:soft_proj}) across the evaluated LoRA pairs and categories, including rank-16 and rank-64 experiments; the parameter controls the trade-off between subject preservation and style transfer (Sec.~\ref{sec:param_study}).
In practice, the projection subspace is constructed using all singular directions of the style LoRA, i.e., we set $k$ equal to the LoRA rank.
Implementation details are provided in the Appendix.
Experiments based on SDXL are performed on an NVIDIA RTX 3090 GPU.

\noindent\textbf{Baselines.}  
We compare our method with representative approaches for combining or learning multiple LoRAs.
Among them, most methods merge two independently trained LoRAs, i.e., the model takes two trained LoRAs as input and produces merged generations.
Specifically, we evaluate 
(i) the direct weight merging baseline $\Delta W_m = \Delta W_c + \Delta W_s$,
(ii) ZipLoRA~\cite{ziplora} (ECCV2024), 
(iii) DuoLoRA~\cite{roy2025duolora} (ICCV2025), 
(iv) K-LoRA~\cite{klora} (CVPR2025), and 
(v) LoRA.rar~\cite{lorarar} (ICCV2025).

\subsection{Asymmetric Projection for Style Preservation}
\label{sec:disc}

\begin{wrapfigure}[15]{r}{0.48\textwidth}
\vspace{-12pt} 
    \centering
    \includegraphics[width=\linewidth, trim=5bp 5bp 5bp 5bp, clip]{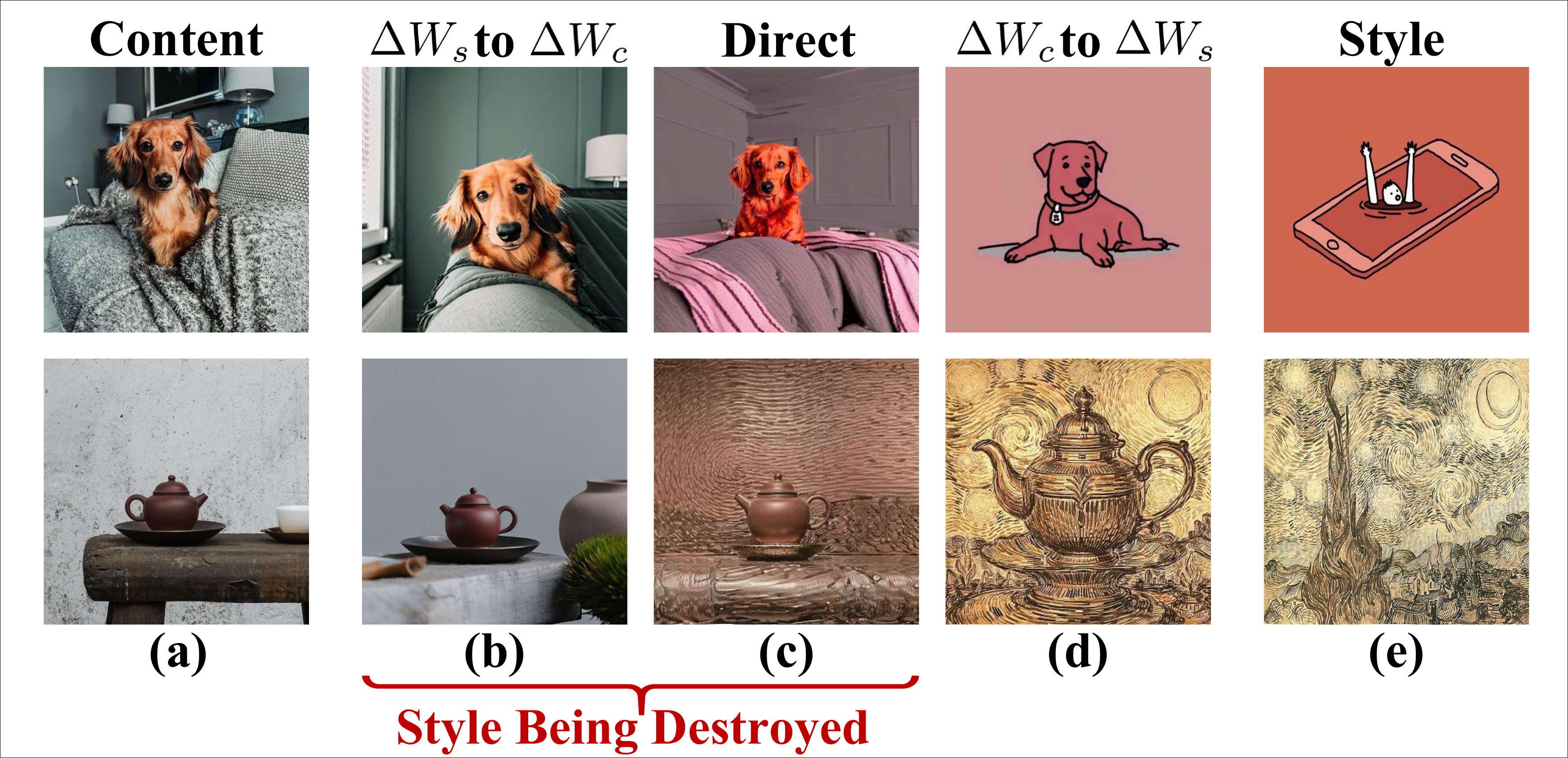}
    \caption{Comparison of LoRA projection. 
    Projecting style into the content null space or direct merging degrades style, while projecting content into the style null space better balances content and style.}
    \label{fig:projection}
\end{wrapfigure}

We first analyze which LoRA should be protected during projection. Direct merging consistently preserves subject structure better than style, suggesting that style representations are more susceptible to cross-LoRA interference. This motivates protecting the style LoRA during fusion.
We compare two projection directions: projecting the style LoRA with respect to the content subspace, and projecting the content LoRA with respect to the dominant style subspace. As shown in Fig.~\ref{fig:projection}, the former substantially weakens stylistic characteristics, while the latter achieves a better content--style balance.
Based on this observation, NP-LoRA adopts an asymmetric projection strategy that preserves style-critical directions while suppressing conflicting content components. This observation is also consistent with Tab.~\ref{tab:main_table} and Fig.~\ref{fig:main_exp}, where direct merging exhibits a clear bias toward content over style.

\subsection{Ablation on Projection Strength} \label{sec:param_study}

\begin{wrapfigure}[15]{r}{0.65\textwidth}
\vspace{-11pt} 
    \centering
    \includegraphics[width=0.95\linewidth, trim=10bp 10bp 10bp 10bp, clip]{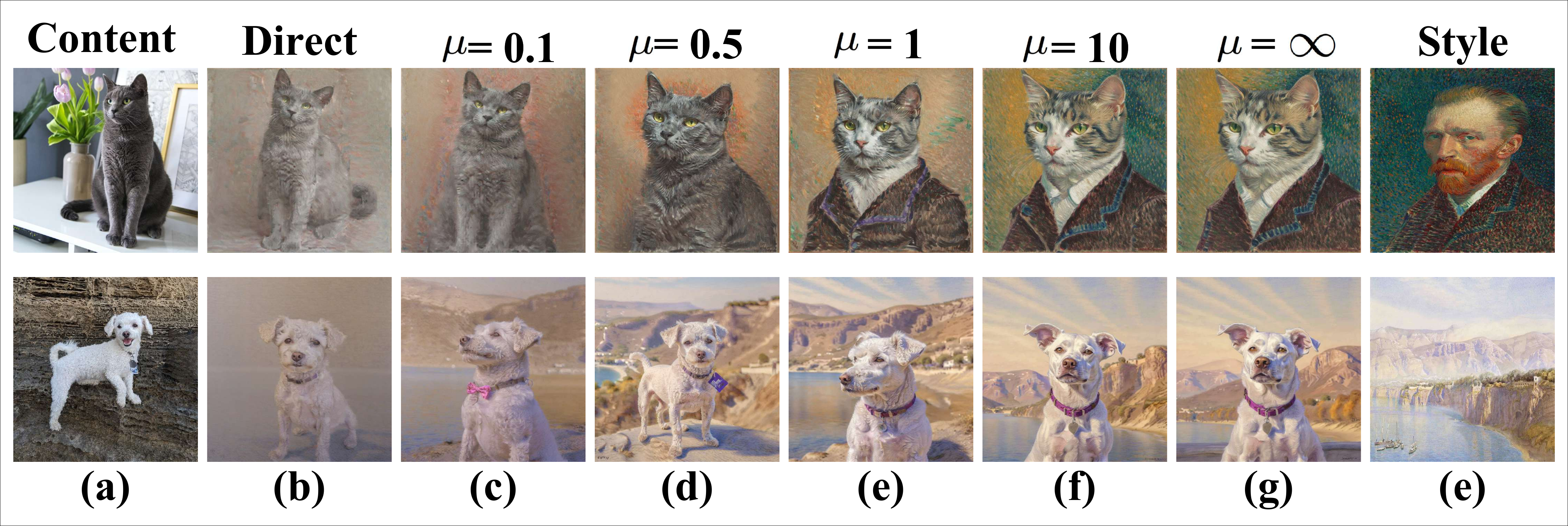}
    \caption{Effect of the projection strength $\mu$ in NP-LoRA.
(a) and (e) are the content and style references, respectively.
(c) corresponds to direct merging ($\mu=0$), while (d)--(g) show increasing projection strengths from soft to hard projection.
Smaller $\mu$ values tend to exhibit stronger content--style interference, whereas larger $\mu$ values better preserve style but may weaken content fidelity.
    }
    \label{fig:mu}
\end{wrapfigure}

We analyze the effect of the projection strength controlled by $\mu$, which adjusts the trade-off between subject fidelity and style consistency, as shown in Fig.~\ref{fig:mu}.
As $\mu$ increases from $0$ to $\infty$, the suppression of content components aligned with the style subspace becomes stronger: $\mu \to 0$ corresponds to direct weight-based merging, while $\mu \to \infty$ approaches hard projection.
We evaluate NP-LoRA with $\mu \in \{0, 0.1, 0.5, 1, 10, \infty\}$ to study this transition.

The results show a clear trade-off.
Smaller $\mu$ values retain more content information but may weaken style consistency, whereas larger $\mu$ values better preserve style but can suppress content details.
In particular, hard projection preserves the style subspace most aggressively, but often loses fine-grained subject information, highlighting the need for a softer projection strategy.

\begin{wraptable}[12]{r}{0.55\textwidth}
\vspace{-6pt} 
    \centering
    \caption{Projection-strength sensitivity on the same 32 LoRA pairs and evaluation protocol as Tab.~\ref{tab:main_table}. Increasing $\mu$ attenuates the content component aligned with the style subspace.}
    \small
    \setlength{\tabcolsep}{1.5pt}
    \begin{tabular}{lcccccc}
    \toprule
    \textbf{Method} & $S^{\text{CLIP}}_{\text{content}}$ & $S^{\text{CLIP}}_{\text{style}}$ & $S^{\text{DINO}}_{\text{content}}$ & $S^{\text{DINO}}_{\text{style}}$ & \textbf{$S_{\text{arith}}$} $\uparrow$ & \textbf{$S_{\text{harm}}$ $\uparrow$} \\
    \midrule
    $\mu=0$ & 0.75 & 0.55 & 0.54 & 0.23 & 0.51 & 0.42 \\
    $\mu=0.1$ & 0.74 & 0.57 & 0.53 & 0.25 & 0.53 & 0.45 \\
    $\mu=0.5$  & 0.73 & 0.59 & 0.52 & 0.33 & \textbf{0.55} & \textbf{0.50} \\
    $\mu=1$ & 0.71 & 0.60 & 0.48 & 0.35 & 0.53 & \textbf{0.50} \\
    $\mu=10$ & 0.69 & 0.60 & 0.40 & 0.37 & 0.52 & 0.48 \\
    $\mu\rightarrow\infty$ & 0.68 & 0.60 & 0.39 & 0.40 & 0.52 & 0.48 \\
    \bottomrule
    \end{tabular}

    \label{tab:mu_sensitivity}
\end{wraptable}
Empirically, $\mu=0.5$ provides the best overall balance and is used as the default setting. The quantitative sweep in Tab.~\ref{tab:mu_sensitivity} shows how the metrics change across projection strengths.


The sweep shows a continuous content--style trade-off: style similarity increases as $\mu$ grows, while content similarity decreases. The default $\mu=0.5$ attains the highest arithmetic mean and ties for the highest harmonic mean among the evaluated settings. For the evaluated pairs, $\mu\in[0.5,1]$ provides a practical starting range for balancing content and style.

\begin{table}[htbp]
\centering
\caption{
Quantitative and preference-based evaluation of content--style LoRA fusion.
(a) Automatic evaluation using CLIP/DINO similarities, with aggregated scores ($S_{\text{arith}}$, $S_{\text{harm}}$) summarizing the overall trade-off.
(b) Preference evaluation by human users and GPT-5, reporting the proportion of cases where each method is selected as the best.
}
\label{tab:main_table}

\begin{minipage}{0.68\linewidth}
\centering
\small
\setlength{\tabcolsep}{1.5pt}
\begin{tabular}{lcccc|cc}
\toprule
\textbf{Method} & $S^{\text{CLIP}}_{\text{content}}$ & $S^{\text{CLIP}}_{\text{style}}$ & $S^{\text{DINO}}_{\text{content}}$ & $S^{\text{DINO}}_{\text{style}}$ & \textbf{$S_{\text{arith}}$} $\uparrow$ & \textbf{$S_{\text{harm}}$ $\uparrow$} \\
\midrule
Direct    & 0.75 & 0.55 & 0.54 & 0.23 & 0.51 & 0.42 \\
ZipLoRA   & 0.75 & 0.56 & 0.55 & 0.24 & 0.52 & 0.44 \\
DuoLoRA & 0.75 & 0.55 & 0.56 & 0.26 & 0.53 & 0.46 \\
K-LoRA    & \textbf{0.76} & 0.55 & \textbf{0.57} & 0.22 & 0.52 & 0.42 \\
LoRA.rar  & 0.70 & 0.53 & 0.50 & 0.30 & 0.51 & 0.46 \\
NP-LoRA   & 0.73 & \textbf{0.59} & 0.52 & \textbf{0.33} & \textbf{0.55} & \textbf{0.50} \\
\bottomrule
\end{tabular}

\vspace{2pt}
 (a) Quantitative similarity metrics
\end{minipage}
\hfill
\begin{minipage}{0.30\linewidth}
\centering
\small
\setlength{\tabcolsep}{3pt}
\begin{tabular}{lcc}
\toprule
Method &  User $\uparrow$ & GPT-5 $\uparrow$ \\
\midrule
Direct     & 8.44\%   & 6.25\%   \\
ZipLoRA    & 9.22\%   & 9.38\%   \\
DuoLoRA    & 11.09\%  & 15.63\%   \\
K-LoRA     & 10.63\%  & 9.38\%    \\
LoRA.rar   & 12.03\%  & 12.5\%    \\
NP-LoRA   & \textbf{48.59\%} & \textbf{46.88\%} \\
\bottomrule
\end{tabular}
\vspace{2pt}
(b) Preference evaluation
\end{minipage}

\end{table}

\subsection{Results}
\label{sec:main_results}
\textbf{Quantitative comparisons.}
Unless otherwise stated, the experiments in this section use SDXL.
Following prior works such as K-LoRA~\cite{klora} (18 pairs) and ZipLoRA~\cite{ziplora} (32 pairs), we evaluate NP-LoRA on 32 randomly selected content--style LoRA pairs, covering 5 animal subjects, 5 object-level categories, and 15 distinct artistic styles.
For each pair, we generate 10 images with shared random seeds across all methods and report results averaged over pairs and seeds.
Unless otherwise specified, all LoRAs are rank-8; results for other ranks are in the appendix.
Unlike idealized settings, our evaluation includes pairs with substantial subspace overlap (often up to 98\%, see Appendix), reflecting realistic entangled LoRA representations.

\begin{figure*}[t]
    \centering
    \includegraphics[width=0.95\linewidth, trim=10bp 10bp 10bp 10bp, clip]{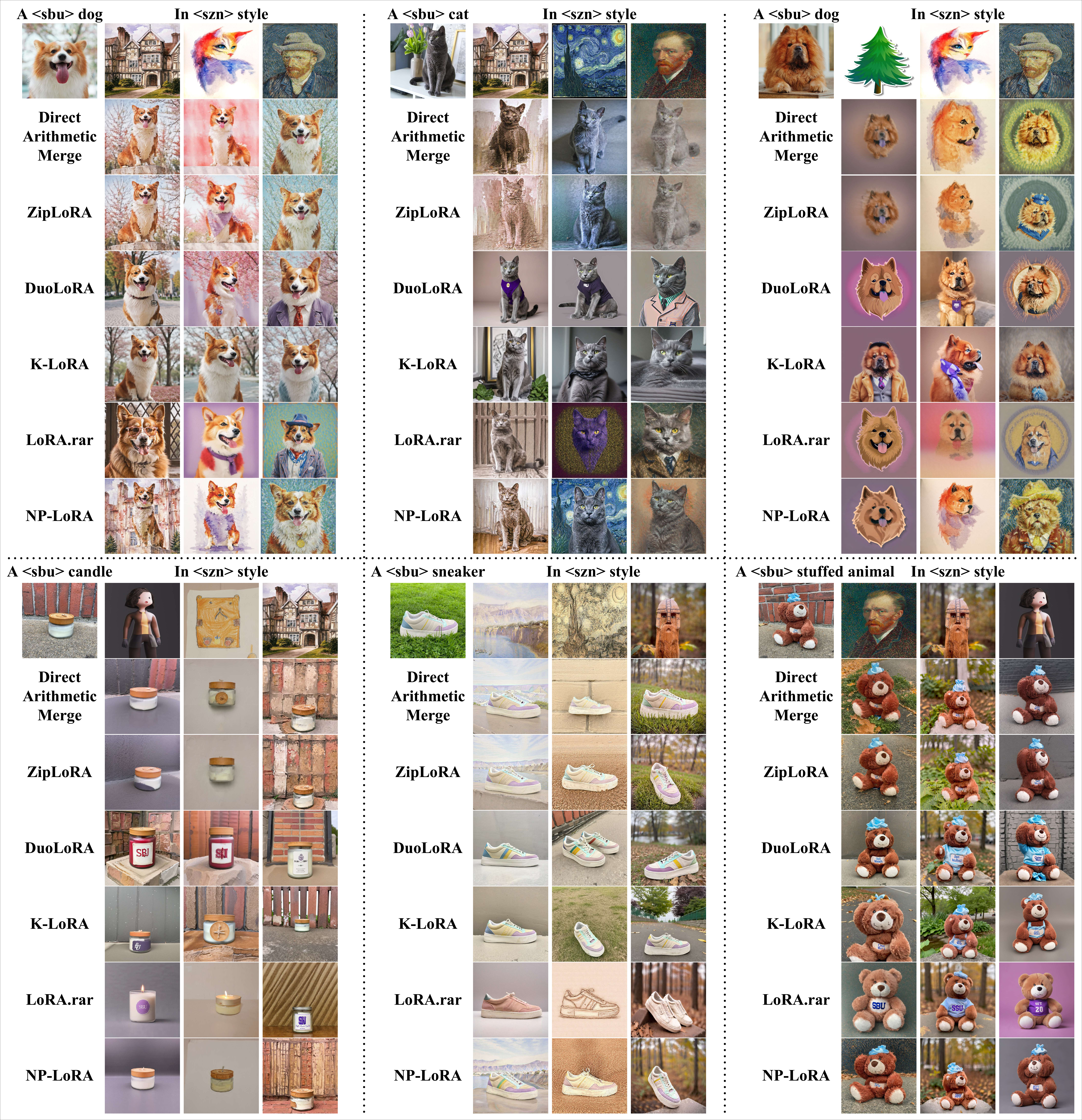}
    \caption{Qualitative comparison with Direct Weight Merge, ZipLoRA, DuoLoRA, K-LoRA, LoRA.rar, and our proposed NP-LoRA, illustrating the trade-off between subject fidelity and style preservation.}
    \label{fig:main_exp}
\end{figure*}

For evaluation, following prior work~\cite{ziplora, klora}, we adopt CLIP~\cite{clip} and DINO~\cite{dino} similarity metrics to jointly assess subject preservation (content) and style alignment (style)~\cite{jiang2024clipdinovisualencoders, ziplora, klora}. 
Content scores compare generated images with subject references, while style scores compare them with style references; CLIP and DINO provide complementary measures of semantic alignment and visual-feature similarity.
Stylization can change texture and local appearance even when subject identity remains recognizable; we therefore interpret the four similarities jointly with human preferences.
Specifically, for CLIP we compute $S^{\text{CLIP}}_{\text{content}}$ and $S^{\text{CLIP}}_{\text{style}}$, 
and for DINO we compute $S^{\text{DINO}}_{\text{content}}$ and $S^{\text{DINO}}_{\text{style}}$. 
To obtain unified content-style similarity measures, we report both the arithmetic mean and the harmonic mean 
following the common F1-style aggregation scheme~\cite{van1979information, f1, ristani2016mtmc, xian2017cvpr}. 
Given the four similarity scores
$S_i \in \{S^{\text{CLIP}}_{\text{content}}, S^{\text{CLIP}}_{\text{style}}, S^{\text{DINO}}_{\text{content}}, S^{\text{DINO}}_{\text{style}}\}$,
we report both the arithmetic mean $S_{\text{arith}}=\tfrac{1}{4}\sum_i S_i$
and the harmonic mean $S_{\text{harm}}=\tfrac{4}{\sum_i (1/S_i)}$, where $S_{\text{arith}}$ reflects the overall trend, while the harmonic mean $S_{\text{harm}}$ emphasizes joint fidelity to both content and style. 
As shown in Tab.~\ref{tab:main_table} (a), our method achieves the highest $S_{\text{arith}}$ and $S_{\text{harm}}$, 
showing its superior ability to balance subject fidelity and style consistency.
Additional rank settings and comparisons with training-based methods are provided in the Appendix. 
In particular, we include results on real-world, community-trained LoRAs, which often exhibit entangled semantics and better reflect practical deployment scenarios.


\begin{wraptable}[14]{r}{0.40\textwidth}
\vspace{-6pt} 
    \centering
    \caption{Efficiency comparison of LoRA merging strategies. Merge includes loading time; Gen./Img. is per-image generation time. LoRA.rar~\cite{lorarar} is excluded for its needs pretraining.}
    \small
    \setlength{\tabcolsep}{5pt}
    \begin{tabular}{lcc}
    \toprule
    Method & Merge (s$^\dagger$)$\downarrow$ & Gen./Img. (s)$\downarrow$ \\
    \midrule
    Direct    & 13.47 & 22.90 \\
    ZipLoRA   & $>10$\,min & 29.93 \\
    DuoLoRA   & $>5$\,min & 23.93 \\
    K-LoRA    & 12.25$^\ast$ & 60.39$^\ast$ \\
    NP-LoRA   & 13.44 & 23.21 \\
    \bottomrule
    \end{tabular}
    {\footnotesize $^\ast$ Only includes LoRA loading time, as K-LoRA performs on-the-fly merging during generation.}

    \label{tab:time_table}
\end{wraptable}

Following K-LoRA~\cite{klora}, we further evaluated perceptual quality through a user study involving 20 participants. The study was conducted on images generated from 32 merged LoRA pairs (Tab.~\ref{tab:main_table} (b)), with 32 images sampled from each pair.
Each case included outputs from SOTA baselines and our method, along with reference subject and style images. Participants selected the result best balancing subject fidelity and style consistency.
As shown in Tab.~\ref{tab:main_table} (b), our method was preferred in $48.59\%$ of cases. 
We further adopt a GPT-5-based automatic evaluation protocol, which is more recent than the MLLM evaluator used in~\cite{lorarar}.
To reduce potential positional and label bias, we randomize the order of candidate images and anonymize method identifiers during evaluation, restoring method names only after collecting the results.
The results show a consistent $46.88\%$ preference in favor of our method.
Across the six evaluated methods, aggregate human and GPT-5 preference rates are strongly correlated (Pearson $r=0.988$; Spearman $\rho=0.928$).

\textbf{Efficiency Comparison.}
To evaluate efficiency, we compare both the merging and generation times in Tab.~\ref{tab:time_table}.
LoRA.rar~\cite{lorarar} is excluded because it depends on pretrained models for parameter fusion. 
ZipLoRA~\cite{ziplora} is the slowest due to additional fusion optimization, whereas K-LoRA~\cite{klora} is efficient during initialization but slows down for dynamic fusion in the sampling stage.
In contrast, NP-LoRA maintains merging and inference efficiency comparable to Direct Merge baseline, yet consistently delivers superior results.

\begin{wrapfigure}[14]{r}{0.7\textwidth}
\vspace{-12pt}
    \centering
    \includegraphics[width=\linewidth]{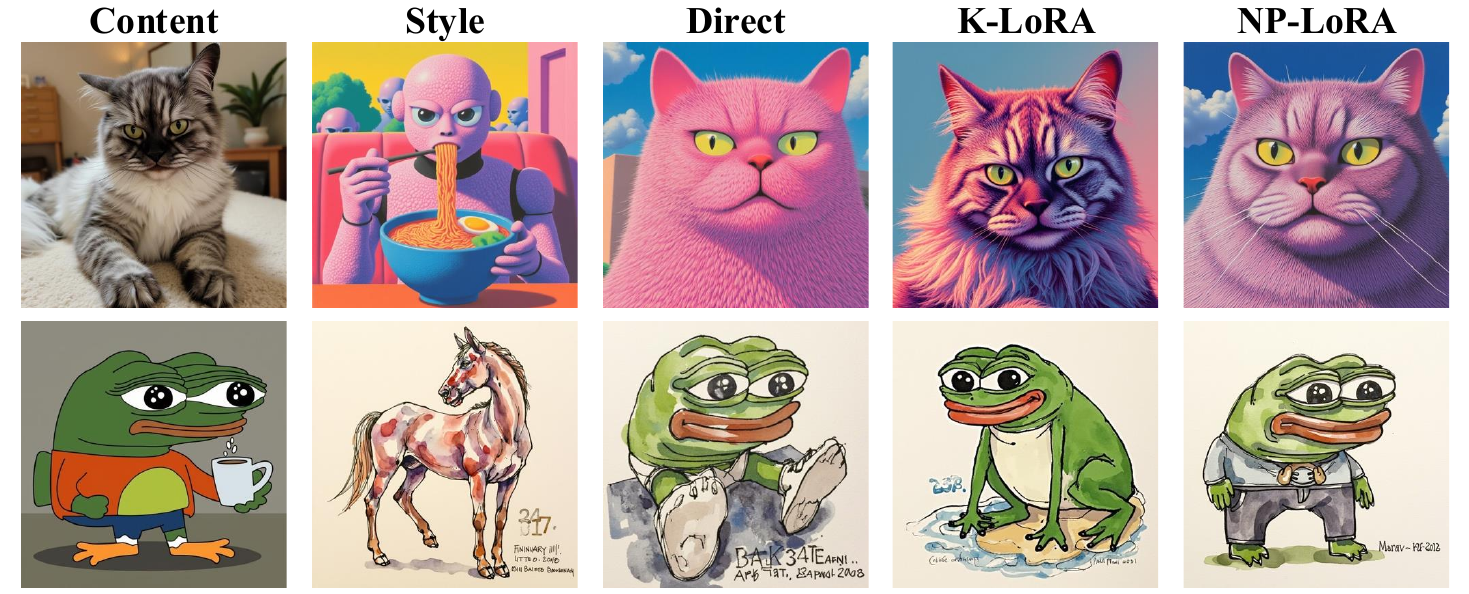}
    \caption{Qualitative comparison on FLUX. Each row shows a subject--style pair and results from Direct Merge, K-LoRA, and NP-LoRA.}
    \label{fig:flux_qual}
\end{wrapfigure}

\textbf{Qualitative comparisons.}
Qualitative results are shown in Fig.~\ref{fig:main_exp}.
Direct merging and existing methods~\cite{ziplora,roy2025duolora,klora,lorarar} achieve only partial fusion, typically preserving either content or style but failing to balance both consistently.
In contrast, NP-LoRA achieves the best results, achieving a superior balance between content fidelity and style preservation.
A qualitative comparison with jointly training method in the Appendix shows that NP-LoRA attains superior results.

\begin{wraptable}[9]{r}{0.60\textwidth}
\vspace{-0pt} 
    \centering
    \caption{Matched quantitative comparison on FLUX using the same LoRA pairs, prompts, and random seeds.}
    \small
    \setlength{\tabcolsep}{1.5pt}
    \begin{tabular}{lcccccc}
    \toprule
    \textbf{Method} & $S^{\text{CLIP}}_{\text{content}}$ & $S^{\text{CLIP}}_{\text{style}}$ & $S^{\text{DINO}}_{\text{content}}$ & $S^{\text{DINO}}_{\text{style}}$ & \textbf{$S_{\text{arith}}$} $\uparrow$ & \textbf{$S_{\text{harm}}$ $\uparrow$} \\
    \midrule
    Direct & 0.71 & 0.55 & 0.59 & 0.36 & 0.55 & 0.52 \\
    K-LoRA & 0.73 & 0.54 & 0.62 & 0.32 & 0.55 & 0.50 \\
    NP-LoRA & 0.70 & 0.57 & 0.58 & 0.40 & \textbf{0.56} & \textbf{0.54} \\
    \bottomrule
    \end{tabular}
    \label{tab:flux_main}
\end{wraptable}

\noindent\textbf{Generalization to FLUX.}

We further evaluate NP-LoRA on the FLUX backbone to examine its applicability beyond SDXL.
We compare it with Direct Merge and K-LoRA~\cite{klora}; the latter provides a public FLUX implementation.
As shown in Fig.~\ref{fig:flux_qual}, all three methods produce recognizable subject--style compositions, while NP-LoRA shows a slightly better balance between subject characteristics and the target style in these examples.
The matched quantitative comparison in Tab.~\ref{tab:flux_main} supports this trend: NP-LoRA attains the highest aggregate scores ($S_{\text{arith}}=0.56$, $S_{\text{harm}}=0.54$) and the highest style similarities among the three methods.


\section{Conclusions}
We propose NP-LoRA, a training-free framework for post-hoc fusion of independently trained content and style LoRAs.
We view LoRA fusion as a subspace interaction problem and use projection to modulate content contributions along dominant style directions.
With a soft projection mechanism, NP-LoRA provides continuous control over the trade-off between subject fidelity and style consistency.
Our analysis and experiments show that content and style LoRAs exhibit substantial subspace interaction, and that controlling these interactions leads to more balanced fusion than weight-based merging.
Extensive evaluations, including quantitative metrics, user studies, and GPT-5-based assessments with randomized and anonymized protocols, show that NP-LoRA consistently improves content-style composition while remaining training-free and computationally efficient.

\noindent\textbf{Limitations.} 
Like prior methods~\cite{ziplora, klora, lorarar}, NP-LoRA treats layers independently, which may leave cross-layer interference unresolved. We evaluate only pairwise subject--style fusion; handling multiple content LoRAs or a joint style subspace remains future work.
\begin{ack}
This study was supported by the National Natural Science Foundation of China (No. 62176252).
The authors declare no competing interests.
\end{ack}

\bibliographystyle{plainnat}   
\bibliography{references}      

\appendix

\section*{Appendix Overview}

This appendix provides additional technical details, derivations, and extended analyses to support the findings presented in the main paper. 
Specifically, we first derive the formulation of the proposed soft projection operator in Sec.~\ref{sec:supp_1} and present details for efficient implementation in Sec.~\ref{sec:supp_2}. 
~Sec.~\ref{sec:supp_3} investigates which null-space construction (U-space or V-space) better preserves stylistic fidelity. 
~Sec.~\ref{sec:supp_4} provides an additional comparison with joint training to further validate the effectiveness of NP-LoRA over training-based approaches. 
~Sec.~\ref{sec:supp_5} demonstrates the controllability of our method under various prompt conditions, while Sec.~\ref{sec:supp_6} evaluates its robustness across different random seeds. 
Sec.~\ref{sec:supp_7} extends our validation to the Flux model to assess cross-model generalization. 
Finally, Sec.~\ref{sec:supp_8} details the GPT-5 evaluation protocol used for quantitative preference assessment.

Together, these sections offer deeper insights into NP-LoRA’s formulation, implementation, and empirical robustness beyond the main paper.

\section{Derivation of the Soft Projection Operator} \label{sec:supp_1}

We start from the relaxed objective introduced in the main text:
\begin{equation}
\min_{\Delta W_c^{\text{proj}}}\; 
\|\Delta W_c^{\text{proj}} - \Delta W_c\|_F^2 
+ \mu\,\|\Delta W_c^{\text{proj}} P\|_F^2,
\label{eq:soft_obj}
\end{equation}
where the first term preserves proximity to the original content adapter, and the second penalizes its energy within the style subspace defined by $P = V_k V_k^\top$.

\paragraph{Closed-form solution.}
Taking the derivative of~\eqref{eq:soft_obj} with respect to $\Delta W_c^{\text{proj}}$ and setting it to zero yields
\begin{equation}
\Delta W_c^{\text{proj}} - \Delta W_c + \mu\,\Delta W_c^{\text{proj}} P = 0,
\end{equation}
which can be rearranged as
\begin{equation}
\Delta W_c^{\text{proj}} (I + \mu P) = \Delta W_c.
\end{equation}
Thus, the solution is given by
\begin{equation}
\Delta W_c^{\text{proj}} = \Delta W_c (I+\mu P)^{-1}.
\label{eq:proj_solution}
\end{equation}

\paragraph{Simplifying the inverse.}
Since $P = V_k V_k^\top$ is a rank-$k$ projector with $V_k^\top V_k = I_k$, the inverse admits the closed form
\begin{equation}
(I+\mu P)^{-1}
= I - \tfrac{\mu}{1+\mu}V_kV_k^\top,
\label{eq:inv_identity}
\end{equation}
which follows from spectral decomposition or the Woodbury identity.

Substituting~\eqref{eq:inv_identity} into~\eqref{eq:proj_solution}, we obtain
\begin{equation}
\Delta W_c^{\text{proj}} 
= \Delta W_c \Big(I - \tfrac{\mu}{1+\mu}V_k V_k^\top\Big).
\end{equation}

\paragraph{Final fusion rule.}
The merged LoRA is then defined as
\begin{equation}
\Delta W_m 
= \Delta W_s + \Delta W_c^{\text{proj}}
= \Delta W_s + \Delta W_c \Big(I - \tfrac{\mu}{1+\mu}V_k V_k^\top\Big).
\end{equation}

This formulation continuously interpolates between direct merging ($\mu = 0$) and hard projection ($\mu \rightarrow \infty$), providing a controllable trade-off between style preservation and content retention.

\section{Randomly Sampling LoRA Pairs.}
Following prior works including K-LoRA~\cite{klora} (18 pairs) and ZipLoRA~\cite{ziplora} (32 pairs), we evaluate our NP-LoRA on \textbf{32 randomly sampled content--style LoRA pairs}, covering \textbf{5 animal subjects}, \textbf{5 object-level categories}, and \textbf{15 distinct artistic styles}.

The specific content--style combinations are listed below:
\begin{center}
\small
\begin{tabular}{p{0.20\linewidth} p{0.70\linewidth}}
\ttfamily
dog &
in watercolor painting style3, in kid crayon drawing style, in oil painting style3, in watercolor painting style2 \\
cat2 &
in watercolor painting style3, in watercolor painting style2, in oil painting style1, in oil painting style2 \\
dog2 &
in sticker style, in watercolor painting style3, in oil painting style3 \\
vase &
in line drawing style, in oil painting style3, in sticker style \\
candle &
in 3d rendering style2, in kid crayon drawing style, in watercolor painting style2 \\
colorful\_sneaker &
in watercolor painting style1, in line drawing style, in wooden sculpture \\
bear\_plushie &
in oil painting style2, in wooden sculpture, in 3d rendering style \\
dog5 &
in cartoon line drawing style, in glowing 3d rendering style \\
wolf\_plushie &
in flat cartoon illustration style, in abstract rainbow colored flowing smoke wave design, in 3d rendering style \\
dog8 &
in cartoon line drawing style, in flat cartoon illustration style, in glowing 3d rendering style, in sticker style \\
\end{tabular}
\end{center}

Content names follow the DreamBooth conventions~\cite{dreambooth}; style captions follow StyleDrop~\cite{styledrop}, with numeric suffixes indicating occurrence order among repeated entries in its \href{https://github.com/aim-uofa/StyleDrop-PyTorch/blob/main/data/data.json}{\texttt{data.json}}.

To further analyze the geometric relationship between independently trained LoRAs, we measure the overlap between two LoRA updates as the proportion of content-LoRA energy lying in the style subspace:
\begin{equation}
\mathrm{Overlap}(A,B)=\frac{\|P_B \Delta W_A\|_F^2}{\|\Delta W_A\|_F^2},
\qquad
P_B = V_B V_B^\top,
\end{equation}
where $V_B$ denotes the orthonormal basis of the right singular subspace of the style LoRA.
Across randomly sampled content--style LoRA pairs, we observe consistently high overlap values, typically around $98\%$, suggesting substantial shared subspace structure between independently trained LoRAs. These overlap values should be interpreted as an empirical indicator of subspace correlation rather than a direct measure of interference strength.

\begin{figure*}[t]
    \centering
    \includegraphics[width=\linewidth, trim=10bp 10bp 10bp 10bp, clip]{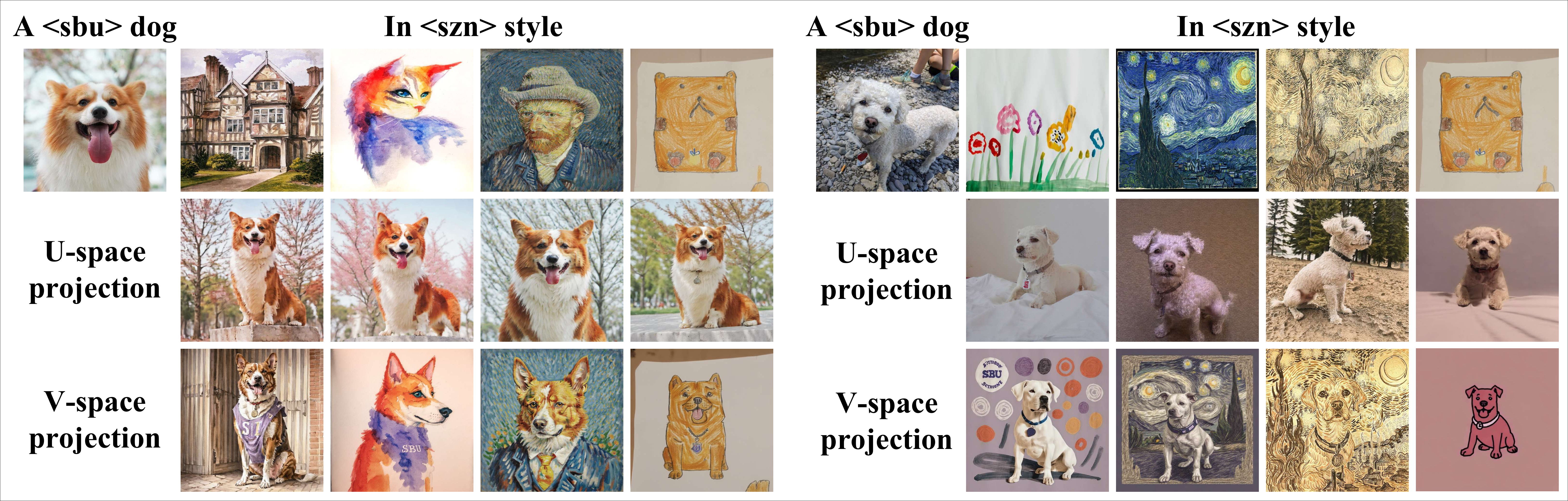}
    \caption{Comparison of output-Space (U) and parameter-space (V) projections for null-space construction.
    The U-space projection fails to remove style interference, leading to distorted stylistic features, 
    whereas the V-space projection effectively eliminates such interference and preserves the intended style appearance.}
    \label{fig:supp_UorV_compressed}
\end{figure*}

\begin{figure*}[t]
    \centering
    \includegraphics[width=\linewidth, trim=10bp 10bp 10bp 10bp, clip]{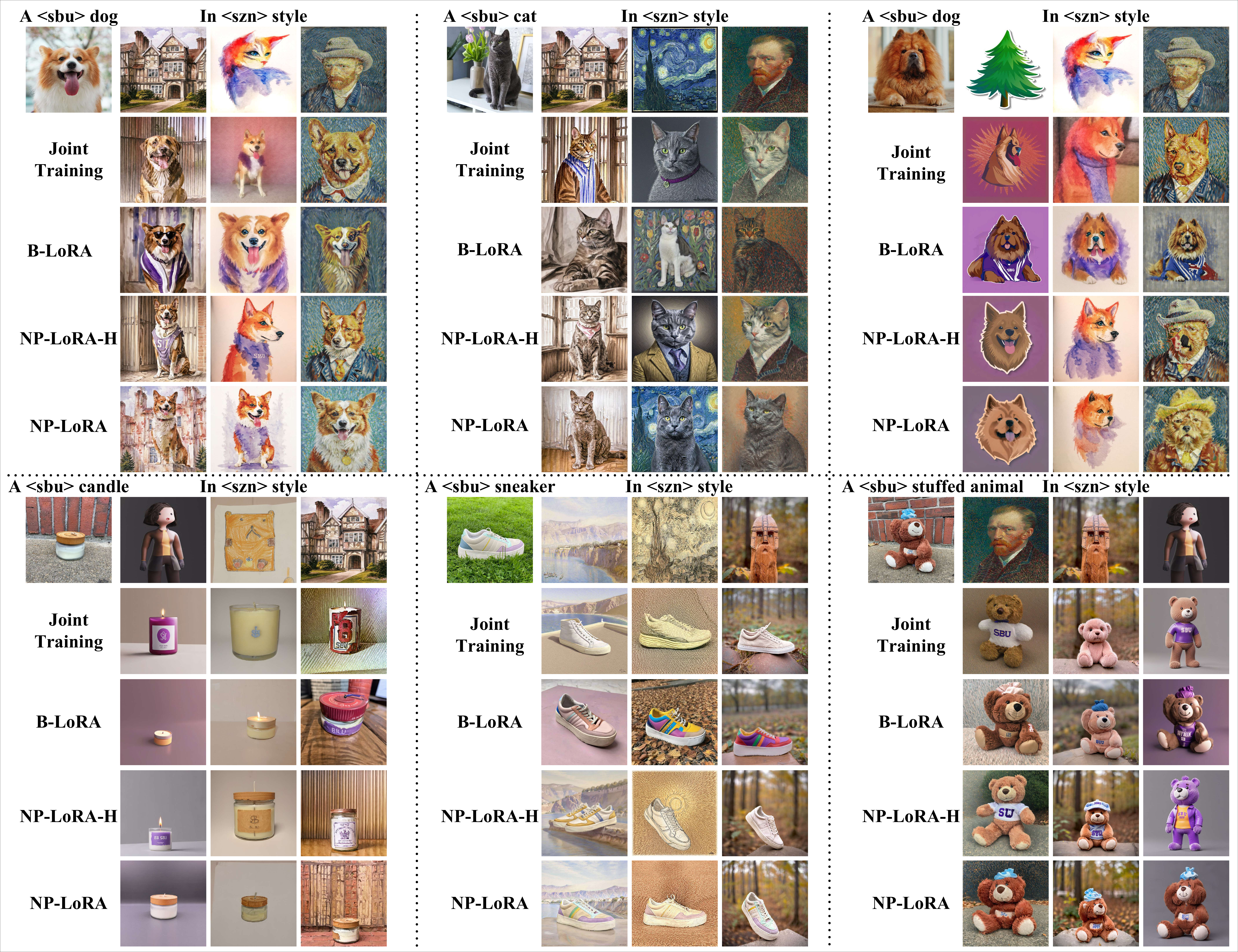}
    \caption{Qualitative comparison with joint training. 
    Joint training exhibits unstable performance and often fails to merge content and style effectively, 
    while N-LoRA achieves consistent and well-balanced fusion without retraining.}
    \label{fig:supp_joint_comparison}
\end{figure*}

\section{Implementation Details} \label{sec:supp_2}

In our theoretical formulation, we employ the singular value decomposition (SVD) to obtain the right singular subspace of the LoRA update matrix 
$\Delta W_s = B_s A_s$, where $A_s \in \mathbb{R}^{r \times n}$ and $B_s \in \mathbb{R}^{m \times r}$. 
Specifically, the original projection of $\Delta W_c$ onto the orthogonal complement of $\Delta W_s$ is computed using the right singular vectors $V_k$ of $\Delta W_s$, i.e.,
\begin{equation}
    \Delta W_c^{\perp} = \Delta W_c (I - V_k V_k^\top).
\end{equation}
However, directly computing $V_k$ via SVD or PCA on the large matrix $\Delta W_s \in \mathbb{R}^{m \times n}$ is computationally expensive, as its complexity scales with $\mathcal{O}(mnr)$.

\begin{figure*}[t]
    \centering
    \includegraphics[width=0.95\linewidth, trim=10bp 10bp 10bp 10bp, clip]{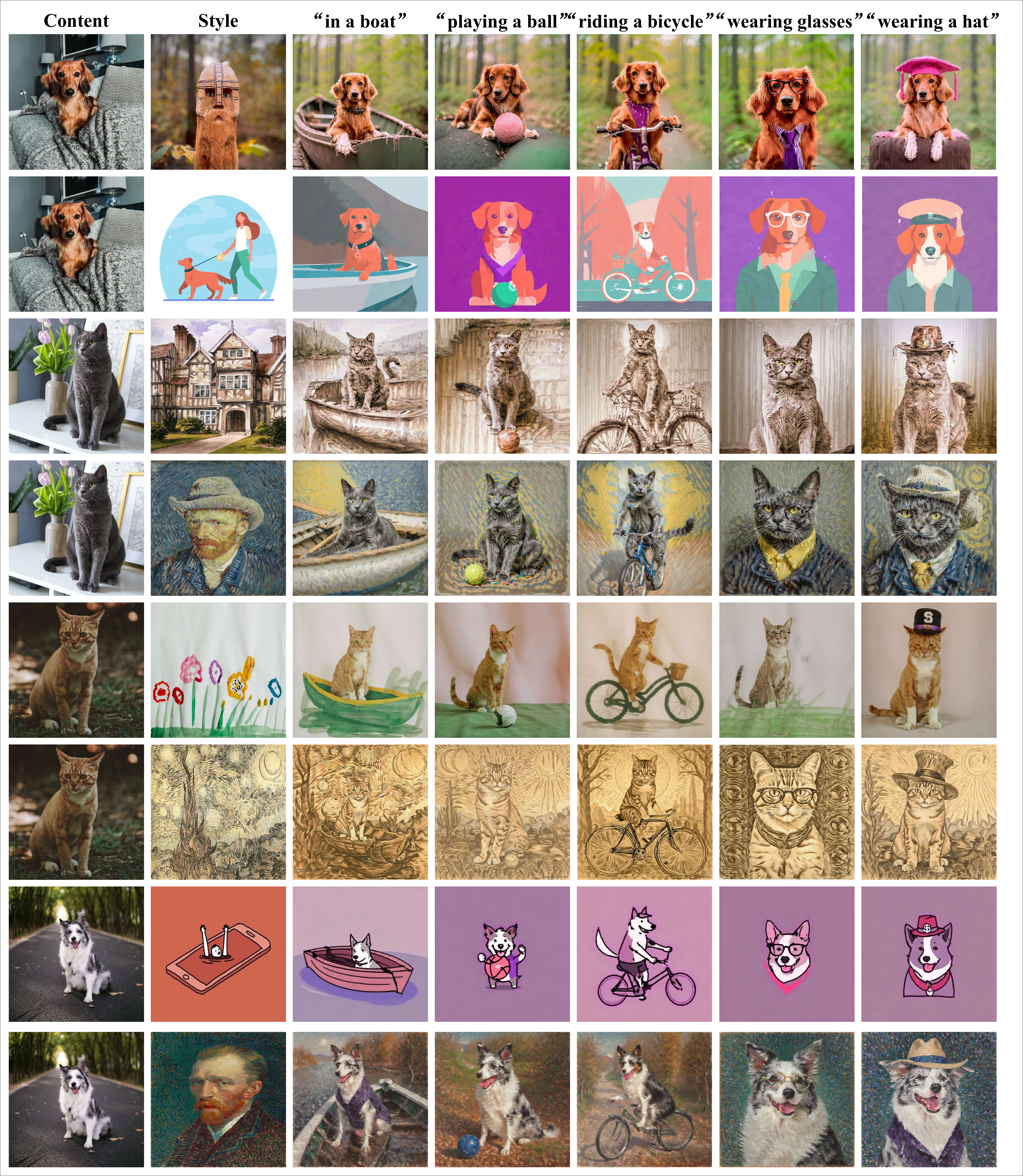}
    \caption{Our method effectively modifies the object’s actions and environment while maintaining the original style.}
    \label{fig:suup_prompt_control_compressed}
\end{figure*}

\paragraph{Efficient QR-based implementation.}
To improve efficiency, we replace the SVD with a QR-based projection that is mathematically equivalent but far more efficient.
Observe that $\Delta W_s$ has rank at most $r$ and its right singular subspace is equal to the column space of $A_s^\top$:
\begin{equation}
    \mathrm{span}(V) = \mathrm{span}(A_s^\top).
\end{equation}
Therefore, we compute a thin QR decomposition on $A_s^\top$:
\begin{equation}
    A_s^\top = Q R,
\end{equation}
where $Q \in \mathbb{R}^{n \times r}$ has orthonormal columns spanning $\mathrm{span}(A_s^\top)$, and $R \in \mathbb{R}^{r \times r}$ is an upper triangular matrix.
We then replace $V$ with $Q$ and compute
\begin{equation}
    \Delta W_c^{\text{proj}} = \Delta W_c - \Delta W_c Q Q^\top,
\end{equation}
which requires only a QR decomposition of a small matrix of size $n \times r$ (where $r \ll n, m$), reducing complexity to $\mathcal{O}(nr^2)$.

\paragraph{Proof of equivalence.}
Let $\Delta W_s = B_s A_s$ denote the LoRA matrix. Since LoRA uses low-rank factors (A$_s \in \mathbb{R}^{r\times n}$) and ($B_s \in \mathbb{R}^{m\times r}$) with full column and row rank (i.e., $rank =r$), ($B_s^\top B_s$) is symmetric positive definite. Then
\begin{equation}
    \Delta W_s^\top \Delta W_s = A_s^\top (B_s^\top B_s) A_s.
\end{equation}
Since $B_s^\top B_s$ is symmetric positive definite, it admits a Cholesky factorization
$B_s^\top B_s = C^\top C$ with $C$ invertible. Substituting gives
\begin{equation}
    \Delta W_s^\top \Delta W_s = (C A_s)^\top (C A_s).
\end{equation}
As $C$ represents a full-rank linear transform, it does not alter the column space of $A_s^\top$.
Hence, the right singular subspace of $\Delta W_s$ coincides with $\mathrm{span}(A_s^\top)$:
\begin{equation}
    \mathrm{Col}(\Delta W_s^\top \Delta W_s) = \mathrm{span}(A_s^\top).
\end{equation}
Therefore, the orthogonal projectors constructed via SVD and QR are identical:
\begin{equation}
    P = V_k V_k^\top = Q Q^\top.
\end{equation}
In practice, this means that replacing the expensive low-rank SVD 
with a simple QR decomposition of $A_s^\top$ yields 
a mathematically equivalent projection while reducing computation time by an order of magnitude.


\section{Which Null-Space Construction Preserves Style: $U$-space or $V$-space?} \label{sec:supp_3}
In the main paper, we construct the null space using the right singular vectors $V$ from the SVD decomposition $\Delta W_s = U \Sigma V^\top$, since interference in LoRA fusion occurs in the parameter space (i.e., the column space of $\Delta W_s$) rather than in the activation space represented by $U$. 
Specifically, $U$ spans the output (activation) domain of $\Delta W_s$, describing how style LoRA influences feature activations, 
while $V$ spans the input (parameter) domain, capturing how LoRA updates are distributed across the weight columns. 
Because LoRA fusion directly manipulates weight parameters, interference emerges along these column directions, making $V$ the appropriate basis for constructing the null space.
Projecting in the $V$-space thus directly eliminates content-induced perturbations along the style-critical directions within the weight domain. 
In contrast, constructing the null space based on $U$ operates in the output (activation) space, which suppresses both content and style responses and weakens overall expressiveness. 
As shown in Fig.~\ref{fig:supp_UorV_compressed}, projections built from $V$ effectively preserve stylistic consistency, while those derived from $U$ fail to disentangle content and style, resulting in noticeable style degradation.

\begin{figure*}[t]
    \centering
    \includegraphics[width=0.85\linewidth, trim=10bp 10bp 10bp 10bp, clip]{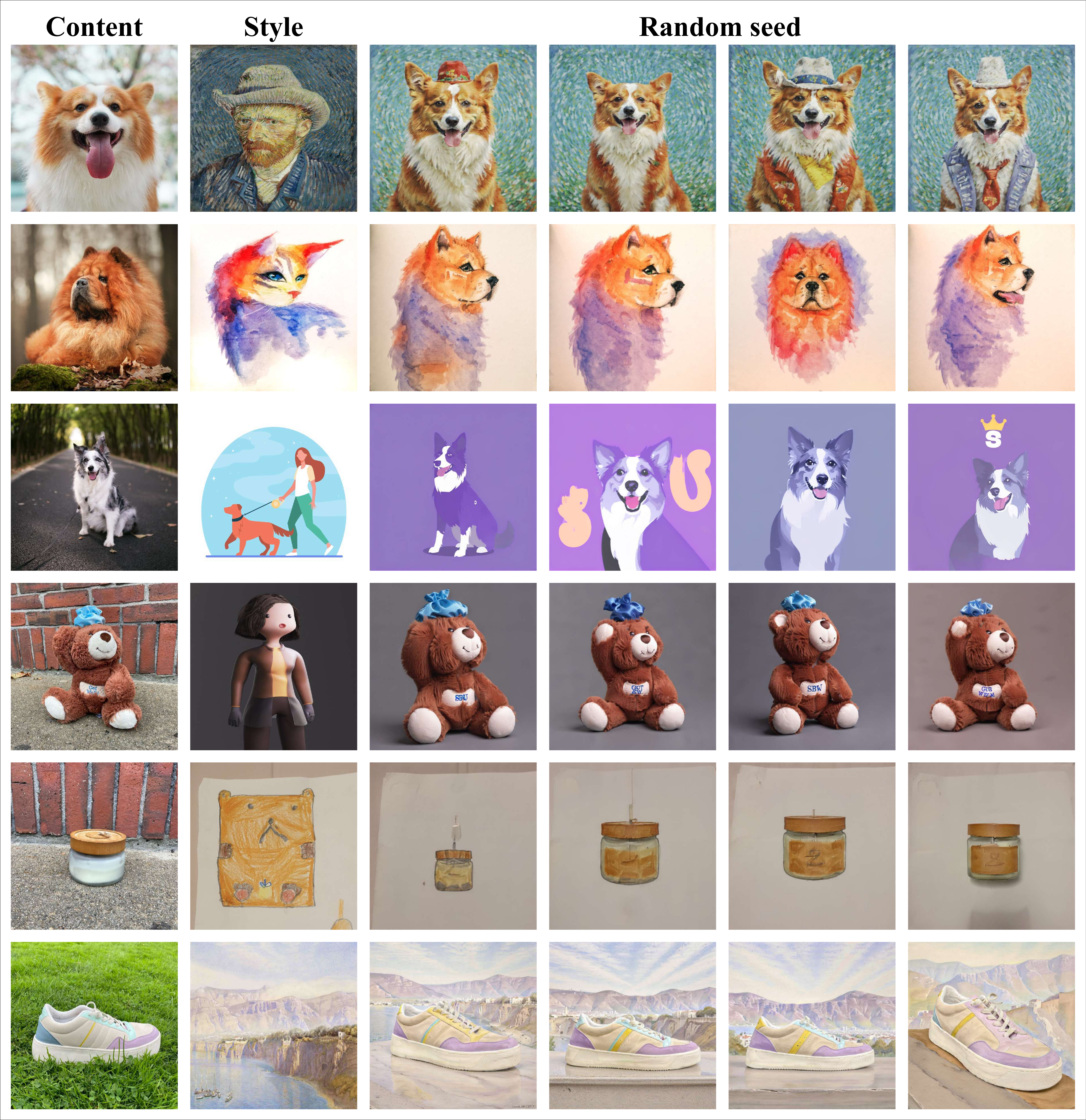}
    \caption{Results obtained with randomly selected seeds demonstrate the stability and robustness of our NP-LoRA.}
    \label{fig:supp_robust_compressed}
\end{figure*}

\section{Additional Ablation on Soft Projection} \label{sec:abl}

We refer to the version without soft relaxation as the hard variant, denoted as NP-LoRA-H.  
The qualitative results are shown in Fig.~\ref{fig:supp_joint_comparison}, and the quantitative comparison is provided in Tab.~\ref{tab:supp_quantitative_joint}.  
As illustrated, NP-LoRA-H effectively preserves the style but tends to lose content details.  
In contrast, our full NP-LoRA achieves a better balance between content and style, yielding the highest overall scores.  
This confirms the necessity of the proposed soft relaxation mechanism.

\section{Generalization to Different LoRA Ranks}
We further conduct the same experiments on LoRAs with different ranks (rank-16 and rank-64). 
As shown in Tab.~\ref{tab:rank_comparison}, the results exhibit consistent trends across both settings, 
leading to the same conclusion: NP-LoRA achieves a better balance between content preservation and style consistency compared to existing methods.

\begin{table}[t]
\centering
\caption{Comparison under different LoRA ranks.}
\label{tab:rank_comparison}

\begin{minipage}{\linewidth}
\centering
\setlength{\tabcolsep}{2pt}
\begin{tabular}{lcccc|cc}
\toprule
\textbf{Method} & $S^{\text{CLIP}}_{\text{content}}$ & $S^{\text{CLIP}}_{\text{style}}$ & $S^{\text{DINO}}_{\text{content}}$ & $S^{\text{DINO}}_{\text{style}}$ & \textbf{$S_{\text{arith}}$} $\uparrow$ & \textbf{$S_{\text{harm}}$ $\uparrow$} \\
\midrule
Direct   & 0.77 & 0.56 & 0.58 & 0.22 & 0.53 & 0.42 \\
ZipLoRA  & 0.79 & 0.57 & 0.59 & 0.23 & 0.54 & 0.44 \\
DuoLoRA  & 0.78 & 0.56 & 0.60 & 0.26 & 0.55 & 0.46 \\
K-LoRA   & 0.80 & 0.55 & 0.60 & 0.22 & 0.54 & 0.43 \\
LoRA.rar & 0.73 & 0.53 & 0.53 & 0.31 & 0.52 & 0.47 \\
NP-LoRA     & 0.78 & 0.60 & 0.58 & 0.30 & 0.56 & 0.50 \\
\bottomrule
\end{tabular}

\vspace{2pt}
\footnotesize (a) Rank 16 LoRAs
\end{minipage}

\vspace{6pt}

\begin{minipage}{\linewidth}
\centering
\setlength{\tabcolsep}{2pt}
\begin{tabular}{lcccc|cc}
\toprule
\textbf{Method} & $S^{\text{CLIP}}_{\text{content}}$ & $S^{\text{CLIP}}_{\text{style}}$ & $S^{\text{DINO}}_{\text{content}}$ & $S^{\text{DINO}}_{\text{style}}$ & \textbf{$S_{\text{arith}}$} $\uparrow$ & \textbf{$S_{\text{harm}}$ $\uparrow$} \\
\midrule
Direct   & 0.80 & 0.55 & 0.61 & 0.23 & 0.55 & 0.44 \\
ZipLoRA  & 0.80 & 0.55 & 0.62 & 0.24 & 0.55 & 0.45 \\
DuoLoRA  & 0.79 & 0.56 & 0.62 & 0.25 & 0.56 & 0.46 \\
K-LoRA   & 0.82 & 0.54 & 0.63 & 0.23 & 0.55 & 0.44 \\
LoRA.rar & 0.73 & 0.54 & 0.55 & 0.31 & 0.53 & 0.48 \\
NP-LoRA     & 0.79 & 0.60 & 0.61 & 0.29 & 0.57 & 0.50 \\
\bottomrule
\end{tabular}

\vspace{2pt}
\footnotesize (b) Rank 64 LoRAs
\end{minipage}

\end{table}

\section{Additional Comparison with Joint Training and B-LoRA} \label{sec:supp_4}
We further compare our method with a joint training baseline, where a new LoRA is trained from scratch using mixed content and style images.
The qualitative results are shown in Fig.~\ref{fig:supp_joint_comparison}, and the quantitative comparison is provided in Tab.~\ref{tab:supp_quantitative_joint}.
As observed, the joint-training approach exhibits high instability: while it occasionally produces reasonable fusion, it often fails to effectively combine content and style.
This instability may arise from catastrophic forgetting during multi-objective optimization, where learning new style features interferes with previously acquired content representations.
We also observe that B-LoRA tends to preserve stylistic patterns but frequently loses content identity, indicating an imbalanced fusion behavior.
In contrast, NP-LoRA achieves consistent and harmonious fusion without retraining, effectively preserving both content fidelity and stylistic characteristics.
Quantitatively, as shown in Tab.~\ref{tab:supp_quantitative_joint}, NP-LoRA outperforms joint training and B-LoRA across all metrics, achieving the highest overall score and demonstrating superior content–style preservation.

\begin{figure*}[t]
    \centering
    \includegraphics[width=0.7\linewidth, trim=10bp 10bp 10bp 10bp, clip]{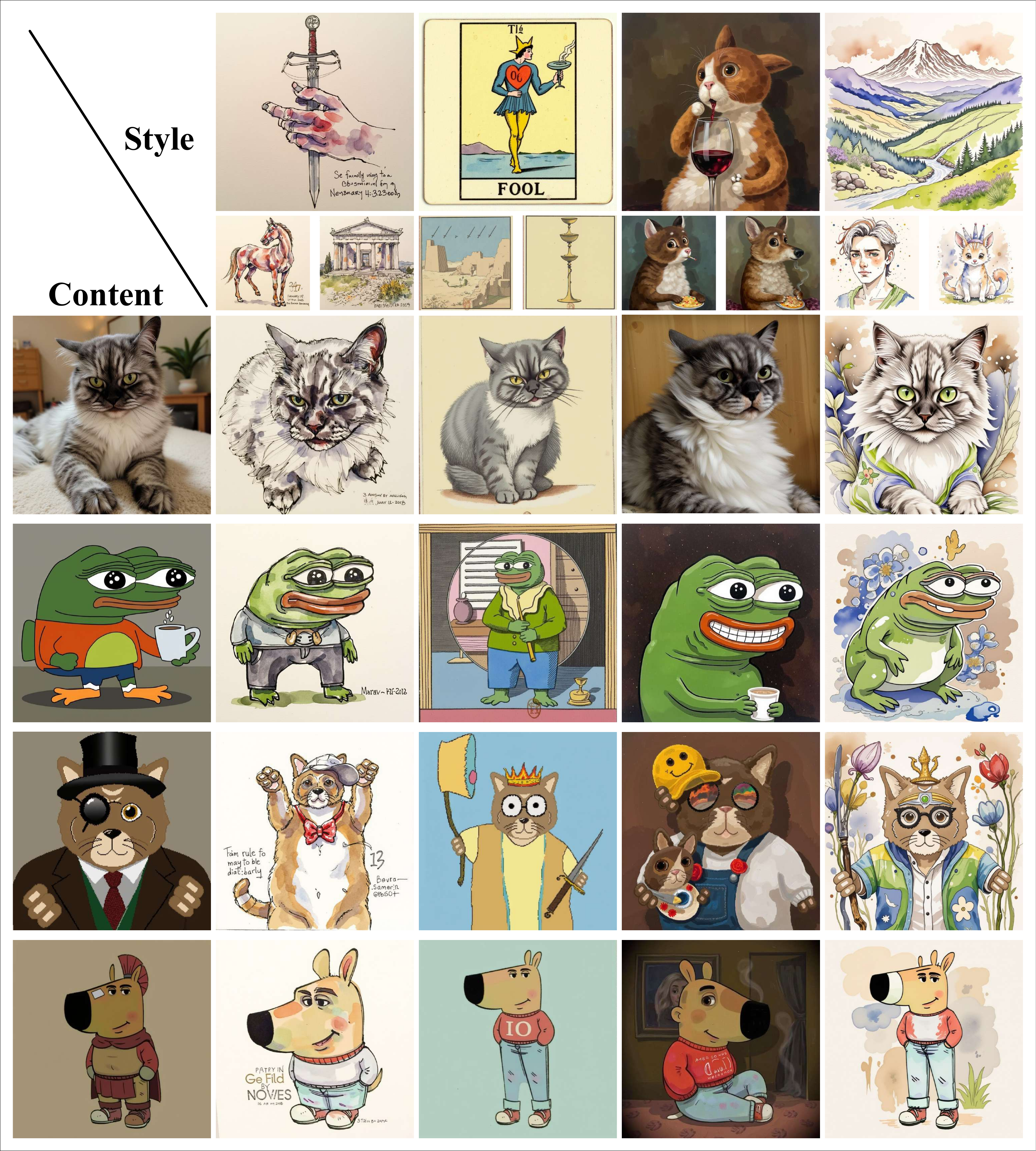}
    \caption{Qualitative results of NP-LoRA on the Flux backbone using diverse publicly available LoRAs.
    Each image corresponds to the combination of the content LoRA shown above and the style LoRA shown on the left, illustrating the results produced by our method.}
    \label{fig:supp_flux_soft_1_compressed}
\end{figure*}

\begin{figure*}[t]
    \centering
    \includegraphics[width=0.7\linewidth, trim=10bp 10bp 10bp 10bp, clip]{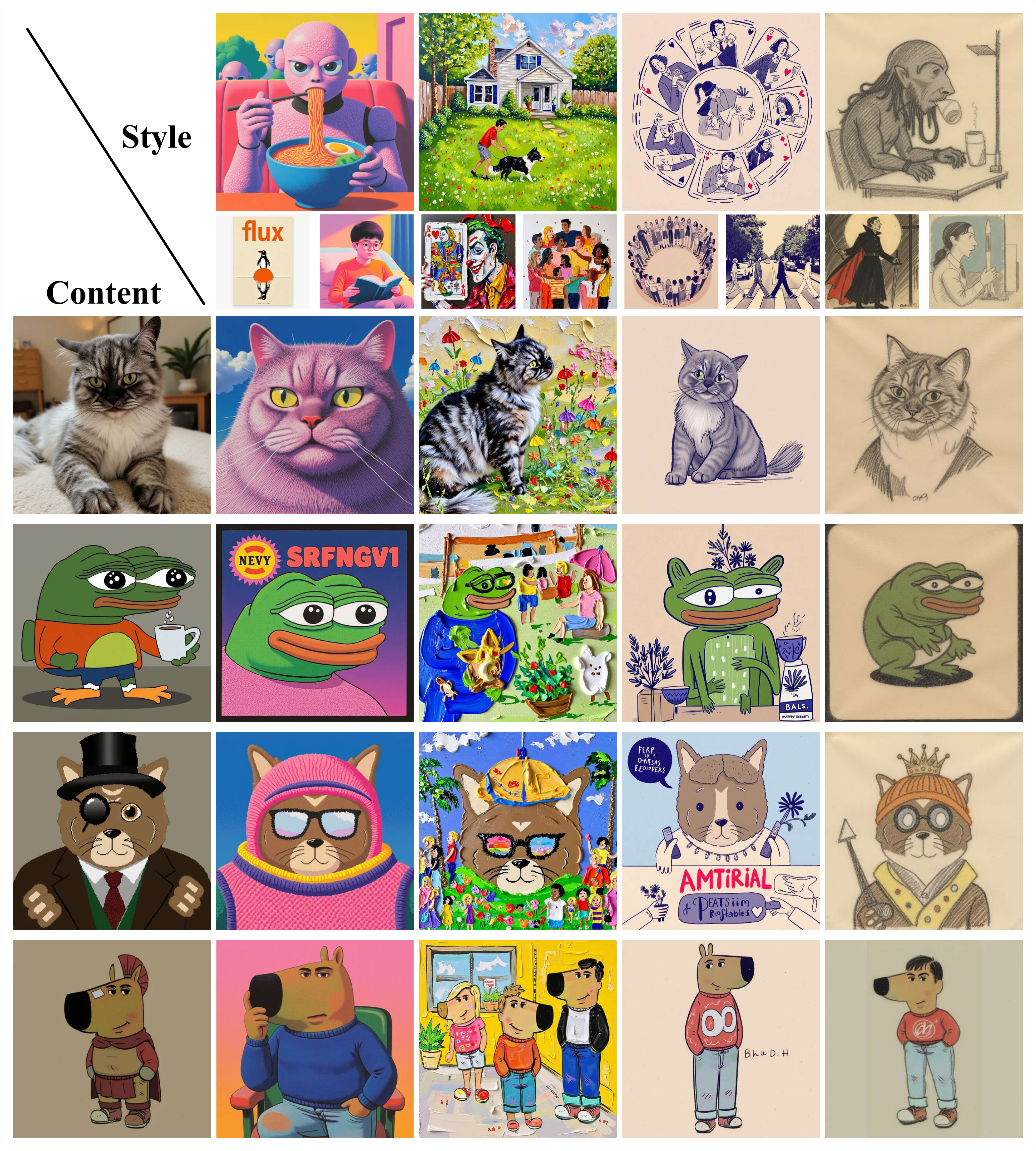}
    \caption{Qualitative results of NP-LoRA on the Flux backbone using diverse publicly available LoRAs.
    Each image corresponds to the combination of the conten t LoRA shown above and the style LoRA shown on the left, illustrating the results produced by our method.}
    \label{fig:supp_flux_soft_2_compressed}
\end{figure*}

\begin{table}[t]
\centering
\caption{
Comparison with joint training methods under the same experimental setting as Tab.~\ref{tab:main_table}.
Evaluation uses CLIP and DINO similarity scores for content preservation and style alignment.
We emphasize the aggregated metrics ($S_{\text{arith}}$ and $S_{\text{harm}}$), as they best reflect the overall trade-off between content and style.
}
\label{tab:supp_quantitative_joint}
\setlength{\tabcolsep}{2pt}
\begin{tabular}{lcccc|cc}
\toprule
\textbf{Method} & $S^{\text{CLIP}}_{\text{content}}$ & $S^{\text{CLIP}}_{\text{style}}$ & $S^{\text{DINO}}_{\text{content}}$ & $S^{\text{DINO}}_{\text{style}}$ & \textbf{$S_{\text{arith}}$} $\uparrow$ & \textbf{$S_{\text{harm}}$ $\uparrow$} \\
\midrule
Joint Training                  & 0.67 & 0.61 & 0.49 & 0.25 & 0.50 & 0.43 \\
B-LoRA     & 0.72 & 0.55 & 0.51 & 0.26 & 0.51 & 0.44 \\
NP-LoRA-H               & 0.68 & 0.60 & 0.39 & 0.40 & 0.52 & 0.48 \\
\textbf{NP-LoRA}      & 0.73 & 0.59 & 0.52 & 0.33 & \textbf{0.55} & \textbf{0.50} \\
\bottomrule
\end{tabular}
\end{table}

\section{Prompt Control} \label{sec:supp_5}
We conduct experiments to assess whether our method can alter the object's actions, the surrounding environment, or introduce new elements by adjusting the prompts. 
As shown in~\ref{fig:suup_prompt_control_compressed}, after modifying the prompts, our method effectively preserves the original object's characteristics and stylistic attributes, while seamlessly integrating new elements or scene details.


\section{Robustness Test} \label{sec:supp_6}
We evaluate the robustness of our method by running experiments with different random seeds to examine stability across stochastic conditions. 
As shown in Fig.~\ref{fig:supp_robust_compressed}, our approach produces consistently stable results under varying seeds, matching the qualitative trends observed in the main paper. 
Specifically, NP-LoRA achieves a balanced trade-off between content preservation and style consistency, yielding the most visually harmonious results.

\section{Further Generalization Validation on FLUX} \label{sec:supp_7}
To further validate the generality of NP-LoRA, we apply our method to the FLUX backbone using publicly available LoRAs covering diverse subjects and styles.
As shown in Fig.~\ref{fig:supp_flux_soft_1_compressed} and Fig.~\ref{fig:supp_flux_soft_2_compressed}, NP-LoRA consistently preserves a better balance between subject identity and stylistic fidelity across various domains.
These results demonstrate that our projection-based fusion mechanism remains effective under different diffusion architectures, highlighting the robustness and general applicability of the proposed approach.

To further evaluate robustness under realistic conditions, we quantitatively compare NP-LoRA on community-trained FLUX LoRAs, where each content--style pair contains 10 generated images.
These LoRAs are publicly available and often contain complex, entangled semantics, making fusion more challenging than cleanly trained LoRAs.
Importantly, NP-LoRA does not require perfectly disentangled LoRAs; instead, it is designed to handle such entangled representations by modulating their interactions.
As shown in Tab.~\ref{tab:flux_quantitative}, NP-LoRA outperforms K-LoRA, the strongest applicable baseline on FLUX, especially in style preservation and overall trade-off.
This confirms that NP-LoRA remains effective for difficult and imperfectly disentangled LoRA pairs, and can robustly handle real-world LoRAs with mixed semantic and stylistic information.

\begin{figure}[t]
    \centering
    \includegraphics[width=0.8\linewidth]{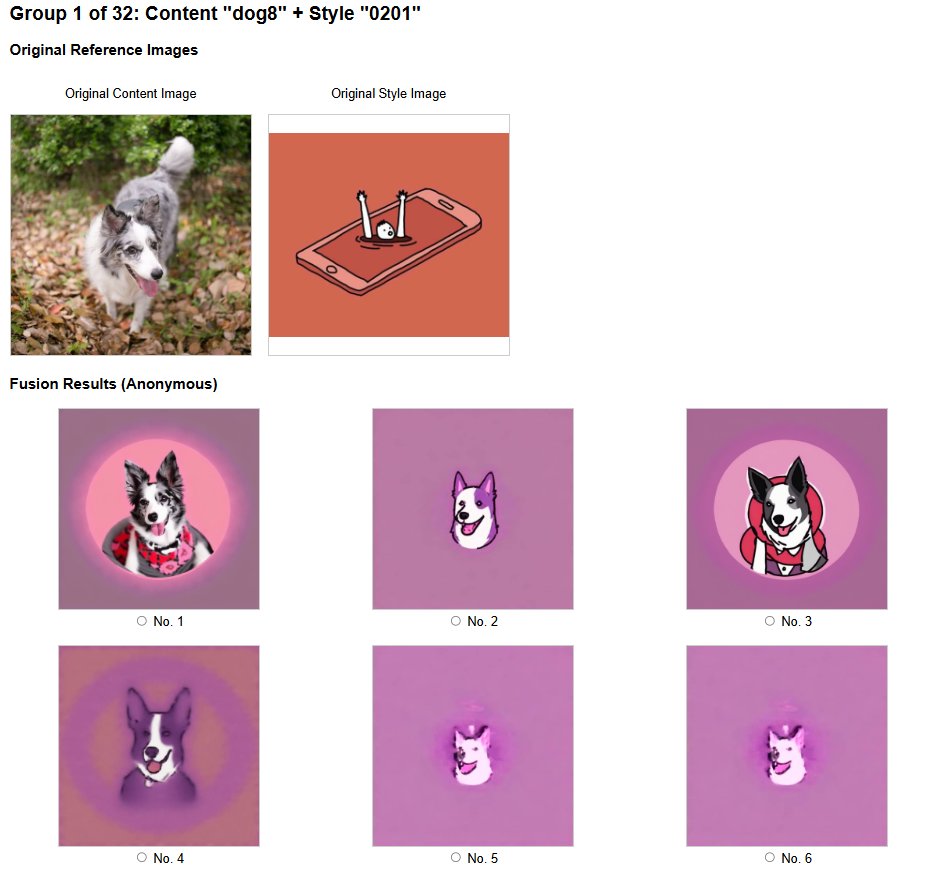}
    \caption{
    Questionnaire screenshot used in the human preference study.
    Participants selected the result that best balanced subject fidelity and style consistency based on the generated outputs and the reference subject/style images.
    }
    \label{fig:placeholder}
\end{figure}

\section{Human Preference Questionnaire}

We provide the questionnaire interface used in the human preference study. The study involved 20 participants and was conducted on images generated from 32 merged LoRA pairs (Tab.~\ref{tab:main_table}(b)), with 32 generated samples evaluated for each pair. For each case, participants were presented with the outputs of NP-LoRA and competing state-of-the-art baselines, together with the corresponding reference subject and style images. Participants were asked to select the result that best balanced subject fidelity and style consistency.
The retained records do not contain per-annotator responses, so inter-annotator agreement cannot be reliably computed.

The evaluation instructions shown to participants were as follows:

\textit{
Evaluation Instructions. Each test group contains the original content image whose subject should be preserved, the original style image whose style should be fused, and fusion results from six anonymous methods. Please select the method that best preserves the content subject while successfully integrating the target style. After submission, your personal selection statistics will be displayed.
}

\begin{table}[t]
\centering
\caption{
Quantitative comparison on community-trained FLUX LoRAs.
NP-LoRA achieves a better overall content--style trade-off than K-LoRA, especially in terms of style preservation and harmonic aggregation.
}
\label{tab:flux_quantitative}
\small
\setlength{\tabcolsep}{4pt}
\begin{tabular}{lcccccc}
\toprule
\textbf{Method} 
& $S^{\text{CLIP}}_{\text{content}}$ 
& $S^{\text{CLIP}}_{\text{style}}$ 
& $S^{\text{DINO}}_{\text{content}}$ 
& $S^{\text{DINO}}_{\text{style}}$ 
& $\mathbf{S_{\text{arith}}}$ 
& $\mathbf{S_{\text{harm}}}$ $\uparrow$ \\
\midrule
K-LoRA  & 0.73 & 0.54 & 0.62 & 0.32 & 0.55 & 0.50 \\
NP-LoRA & 0.70 & 0.57 & 0.58 & 0.40 & 0.56 & 0.54 \\
\bottomrule
\end{tabular}
\end{table}

\paragraph{Prompt used for GPT-5 evaluation.}
\begin{quote}
\small\ttfamily
Task: Evaluate and select the candidate image that best balances Subject Fidelity and Style Consistency.

Inputs:
- First image = A.jpg: Subject Fidelity reference. It shows the core subject/content that should be preserved.
- Second image = B.jpg: Style reference. It shows the artistic style, colors, texture, and lighting that should be applied.
- The remaining six images are candidate results.
- Candidate images are anonymized and shown in randomized order.
- Each candidate is labeled only by an anonymous ID: $Candidate_1$, $Candidate_2$, $Candidate_3$, $Candidate_4$, $Candidate_5$, $Candidate_6$.

Evaluation Criteria:
1) Subject Fidelity: retain the key characteristics, structure, and recognizability of the subject in A.jpg.
2) Style Consistency: apply the style, texture, colors, and lighting of B.jpg.
3) Balance: prefer the candidate that best preserves the subject while also matching the reference style.

Crucial Requirement:
Do not choose an image that only preserves the subject but fails to apply the style.
Do not choose an image that strongly mimics the style but distorts or loses the subject.
Select the candidate with the best overall balance between subject fidelity and style consistency.

Output Rules:
You MUST respond with ONLY ONE of the following exact strings:

$Candidate_1$
$Candidate_2$
$Candidate_3$
$Candidate_4$
$Candidate_5$
$Candidate_6$

Do NOT output anything else.
Do NOT include punctuation, brackets, markdown, or explanation.
\end{quote}

\section{Details for GPT-5 Evaluation Protocol}
\label{sec:supp_8}

In the main paper, we employed GPT-5 to automatically evaluate and select the candidate image that best balances Subject Fidelity and Style Consistency. 
To reduce potential positional and label bias, we randomized the order of candidate images and anonymized all method identifiers during evaluation. 
The exact prompt used for GPT-5 is listed below to ensure full reproducibility.

For each evaluation case, GPT-5 received the reference subject image (\texttt{A.jpg}), the reference style image (\texttt{B.jpg}), and six anonymized candidate images labeled only as \texttt{Candidate\_1}--\texttt{Candidate\_6}. 
The mapping between anonymous candidate labels and actual methods was randomly generated for each case and restored only after evaluation for result aggregation. 
Each case was judged independently to avoid cross-sample bias. 
The final GPT-5 preference score reports the proportion of cases in which each method was selected as the best-balanced result.

\section{Broader Impact}

This work studies training-free LoRA fusion for controllable image generation, which may positively impact efficient personalization and creative content generation by enabling flexible composition of subject and style adapters without additional training.

At the same time, improved controllable generation techniques may also introduce potential misuse risks, including unauthorized identity/style composition, misleading synthetic media generation, or misuse of publicly shared LoRA adapters. Our work does not introduce a new generative backbone model or dataset, but instead focuses on improving the composition of existing LoRA adapters. We encourage responsible use consistent with the licenses and intended purposes of the underlying models and assets.




\end{document}